\documentclass[10pt,twocolumn,letterpaper]{article}

\usepackage[accsupp]{axessibility}  % Improves PDF readability for those with disabilities.

\usepackage{iccv}
\usepackage{times}
\usepackage{epsfig}
\usepackage{graphicx}
\usepackage{amsmath}
\usepackage{amssymb}
\usepackage{listings}

\usepackage{paralist}
\usepackage{float}
\usepackage{longtable}
\usepackage{soul}

\usepackage{booktabs}
\usepackage[dvipsnames]{xcolor}
\usepackage{pifont} % cmark/xmark

\usepackage[breaklinks=true,colorlinks,bookmarks=false]{hyperref}

\iccvfinalcopy % *** Uncomment this line for the final submission

\def\vlfull{{Vision \& Language}}
\def\vl{{VL}}
\def\vlcfull{{Visual Language Concepts}}
\def\vlc{{VLC}}
\def\vlchecklist{{VL-Checklist}}
\def\winoground{{Winoground}}
\def\ARO{{ARO}}
\def\ourdataset{{\textbf{\blue{SyViC}}}} % just to make a distinction from model-generated synthetic data vs. Simulated  Synthetic data?
\def\ourdatasetfull{{\textbf{\blue{Sy}}nthetic \textbf{\blue{Vi}}sual \textbf{\blue{C}}oncepts}}
\def\ours{{syn-}}

\newcommand{\cmark}{\textcolor{blue}{\ding{51}}}%
\newcommand{\xmark}{\ding{55}}%

\newcommand\blue[1]{\textcolor{blue}{#1}}
\newcommand\red[1]{\textcolor{red}{#1}}
\newcommand\green[1]{\textcolor{citecolor!80}{#1}}
\definecolor{cyan}{rgb}{0.45,0.87,0.95}
\definecolor{orange}{rgb}{0.95,0.8,0.6}
\definecolor{britishracinggreen}{rgb}{0.0, 0.26, 0.15}
\definecolor{cadmiumgreen}{rgb}{0.0, 0.42, 0.24}

\newcommand{\gcol}[1]{{\bf \fontsize{5.5}{42}\selectfont \color{citecolor!80}~(#1)}}
\newcommand{\rcol}[1]{{\bf \fontsize{5.5}{42}\selectfont \color{lightred!180}~(#1)}}
\definecolor{citecolor}{RGB}{34,139,34}
\definecolor{lightred}{RGB}{241,140,142}

\begin{document}

%%%%%%%%% TITLE
\title{Going Beyond Nouns With Vision \& Language Models Using Synthetic Data}

\author{
    \textbf{Paola Cascante-Bonilla\thanks{Equal contribution. Project page: \href{https://synthetic-vic.github.io/}{https://synthetic-vic.github.io/}}\,\,\thanks{Work partially done while interning at the MIT-IBM Watson AI Lab.}\,\,$^{1,2}$
    \quad Khaled Shehada\textsuperscript{$*$}$^{2,3}$
    \quad James Seale Smith$^{2,4}$
    \quad Sivan Doveh$^{6,7}$} \\
    \textbf{Donghyun Kim$^{2,7}$
    \quad Rameswar Panda$^{2,7}$
    \quad Gül Varol$^{5}$
    \quad Aude Oliva$^{2,3}$} \\
    \textbf{Vicente Ordonez$^{1}$
    \quad Rogerio Feris$^{2,7}$
    \quad Leonid Karlinsky$^{2,7}$} \\
    \normalsize
    $^{1}$Rice University\quad $^{2}$MIT-IBM Watson AI Lab\quad $^{3}$MIT
    \quad $^{4}$Georgia Institute of Technology\\
    \normalsize
    $^{5}$LIGM, \'{E}cole des Ponts\quad$^{6}$ Weizmann Institute of Science \quad$^{7}$IBM Research\\
}

\maketitle
% \thispagestyle{empty}

%%%%%%%%% BODY TEXT
%%%%%%%%% ABSTRACT
\begin{abstract}
Large-scale pre-trained Vision \& Language (\vl{}) models have shown remarkable performance in many applications, enabling replacing a fixed set of supported classes with zero-shot open vocabulary reasoning over (almost arbitrary) natural language prompts. However, recent works have uncovered a fundamental weakness of these models. For example, their difficulty to understand \vlcfull{} (\vlc{}) that go `beyond nouns' such as the meaning of non-object words (e.g., attributes, actions, relations, states, etc.), or difficulty in performing compositional reasoning such as understanding the significance of the order of the words in a sentence. In this work, we investigate to which extent purely synthetic data could be leveraged to teach these models to overcome such shortcomings without compromising their zero-shot capabilities. 
We contribute \ourdatasetfull{} (\ourdataset{}) - a million-scale synthetic dataset and data generation codebase allowing to generate additional suitable data to improve \vlc{} understanding and compositional reasoning of \vl{} models. Additionally, we propose a general \vl{} finetuning strategy for effectively leveraging \ourdataset{} towards achieving these improvements.
Our extensive experiments and ablations on \vlchecklist{}, \winoground{}, and \ARO{} benchmarks demonstrate that it is possible to adapt strong pre-trained VL models with synthetic data significantly enhancing their VLC understanding (e.g. by 9.9\% on \ARO{} and 4.3\% on \vlchecklist{}) with under $1\%$ drop in their zero-shot accuracy.
\end{abstract}
\vspace{-0.4cm}
\section{Introduction}
\label{sec:intro}
\begin{figure}[t!]
    % \vspace{-0.5cm}
    \centering
    \includegraphics[width=\linewidth]{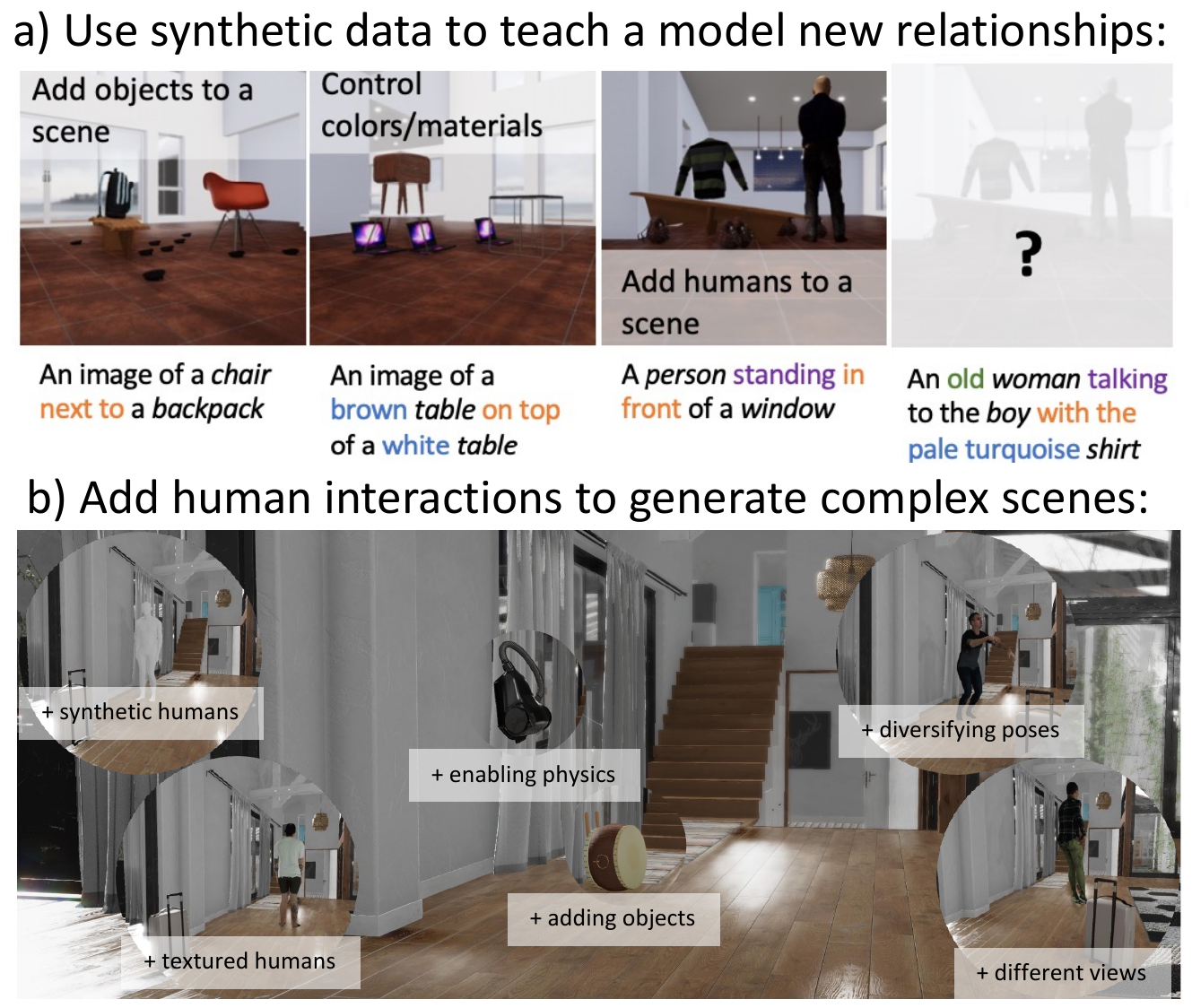}
    \caption{Overview of our proposed synthetic set: we place different objects in a scene and change their position, color, size, and material. We further emphasize on human-level interactions, sampling a wide set of body poses and behaviors to cover transitive and intransitive human actions. } % (e.g., running and throwing). }.
    \label{fig:figure1}
    \vspace{-0.25cm}
\end{figure}

There have been impressive advances in the performance of zero-shot recognition through the use of large-scale pre-trained \vlfull{} (\vl{}) models~\cite{clip,align,singh2022flava,blip,cyclip,filip,declip,pyramidclip}. However, these \vl{} models still face some important challenges in understanding \vlcfull{} (\vlc{}) beyond object nouns (e.g.,~recognizing attributes, relations, states) and in terms of compositional reasoning capabilities (i.e..,~understanding subtle changes in meaning due to small changes in word order). 
Recently, several benchmark tests have been devised to demonstrate the extent to which these models lack these capabilities~\cite{winoground,vlc,aro}
\footnote{Please also see supplementary material for the expanded set of results of~\cite{vlc} including results for all the most recent open-sourced \vl{} models, all exhibiting poor \vlc{} understanding performance.}. 
% \footnote{Check Appendix for the expanded set of results of~\cite{vlc} including results for all the most recent open-sourced \vl{} models, all exhibiting poor \vlc{} understanding performance.}.
As noted in several recent works~\cite{aro,vlc,ours_teaching}, this behavior of \vl{} models is likely due to the contrastive pre-training prevalent for all of them and likely inducing `bag-of-objects' kind of representations (for both images and text alike). Indeed, for (even large) random batches of paired image-text samples, the collection of objects (nouns) in the image (or text) is likely to uniquely determine the image (or text) in the batch, making contrastive batch losses focus on the objects (nouns) while regarding other details (attributes, relations, states, word order, etc.) as unnecessary. Intuitively, this impairs \vlc{} understanding and compositional reasoning of the resulting model.

Given the above, a natural question to ask is what is the most effective way to `fix' the \vl{} models to improve their \vlc{} understanding and compositional reasoning performance? An approach proposed in concurrent works~\cite{ours_teaching,aro}, advocates for the use of text augmentation, using language tools to teach a model the importance of non-noun words by manipulating them (e.g., replacing them with incorrect alternatives) and adding the resulting texts to the same batch. Although effective, such augmentation techniques are only easy on the text side and are much harder and prohibitively expensive on the image side. Indeed, finding, collecting, or generating real image samples sharing the objects but differing in their composition, attributes, relations, or states is very difficult. Although significant progress has been achieved with text-based editing~\cite{hertz2022prompt,kawar2022imagic,mokady2022null,brooks2022instructpix2pix,kim2022diffusionclip,avrahami2022blended,parmar2023zero}, these methods are relatively slow (leveraging diffusion) and not sufficiently stable to allow effective use for augmentation in training pipelines. In this work, therefore, we propose an orthogonal route -- \vl{}
data synthesis for fixing \vl{} models by targeted demonstration. Specifically, we propose enhancing the \vlc{} and compositionality aspects of the generated \textit{visual and text} data, in turn using this data for finetuning \vl{} models teaching them to pay closer attention to these aspects. Moreover, besides being largely free and infinitely scalable, synthetic data has an additional advantage -- it can also be free from privacy concerns always accompanying real data. 

Besides the inherent challenges of realistic data simulation, building synthetic data that can be effectively used to improve \vlc{} and compositionality aspects of \vl{} models pre-trained on massive real data poses additional technical challenges. 
Unlike the majority of prior work focusing on synthetic visual data generation, we need not only to generate images, but also the text that describes compositional items in a scene. 
We generate synthetic videos that leverage realistic physical 3D simulation~\cite{TDW} including diverse 3D environments and different 3D objects, human motions, and actions assets~\cite{athanasiou22teach, AMASS,petrovich2022temos, BABEL}, added interaction with objects, and different camera viewpoints. 
Every frame of these videos is accompanied by rich metadata, allowing using language grammar for generating detailed descriptive captions of any instantaneous scene in each video. These captions, in turn, allow collecting diverse image-text pairs samples contrasting which one to another highlights to the model the importance of the compositional items in the text captions (e.g. different viewpoints or different frames in the same video share objects but may strongly differ in the \vlc{} and other compositional items). 
While motion assets were used by previous works to generate synthetic data~\cite{varol17_surreal,varol21_surreact}, the visual data was not accompanied by textual captions and was not designed with the need to highlight compositionality in mind.
We contribute \ourdatasetfull{} (\ourdataset{}) -- a large (million-scale) generated synthetic \vl{} dataset with rich textual captions, easily extensible through our data synthesis code 
% that is provided in supplementary and would be released upon acceptance 
together with all the already generated million-scale synthetic data used in this paper (Figure~\ref{fig:figure1}).

In addition to the data synthesis pipeline, we also offer a strategy for effectively leveraging the generated synthetic data, while avoiding forgetting real data alignment and losing the strong a-priori zero-shot capabilities of the model. We propose and extensively ablate a combination of domain adaptation by stylization~\cite{zhao2022style}, parameter efficient fine-tuning~\cite{lora}, long captions handling, and model averaging methods~\cite{wortsman2022robust} to reduce forgetting, as well as examine the effect of different aspects of data synthesis and finetuning choices on the gains in \vlc{} and compositionality understanding.

Our contributions can be summarized as follows: (i) we contribute \ourdataset{} -- a million-scale synthetic dataset with rich textual captions, intended for improving \vlc{} understanding and compositional reasoning in \vl{} models, as well as the methodology and the generation codebase
\footnote{We release our code together with all million-scale synthetic data used in this paper here: \href{https://github.com/uvavision/SyViC}{https://github.com/uvavision/SyViC}}\,
for its synthesis and potentially extensibility; (ii) an effective general \vl{} model finetuning strategy
enabling effective leveraging of \ourdataset{} data for enhancing the aforementioned aspects of strong pre-trained \vl{} models without sacrificing their zero-shot capabilities; (iii) experimental results and extensive ablation study showing significant (over 10\% in some cases) improvement in \vlc{} understanding and compositional reasoning respectively, measured on all the recent \vlchecklist{}, \ARO{}, and \winoground{}  benchmarks and validated on the most popular CLIP~\cite{clip} model and its derivatives (e.g. the most recent CyCLIP~\cite{cyclip}).

For supplemental materials, readers are referred to the associated arXiv document at [arXiv:2303.17590].
\section{Related Work}
\label{sec:relatedWork}

\noindent{\textbf{Large-scale Vision\&Language (VL) Models:}}
Large-scale pre-trained VL models such as CLIP~\cite{clip} or ALIGN~\cite{align} show remarkable success in many zero-shot recognition tasks such as image classification or detection~\cite{zareian2021open}. Despite the continued advancements made in this direction~\cite{cyclip,filip,declip,pyramidclip}, recent studies (\eg,~\cite{vlc,winoground,aro}) show that existing VL models exhibit limited comprehension of structured vision language concepts (VLC). Yuksekgonul~\etal~\cite{aro} argue that contrastive learning for image-retrieval learns shortcuts and does not learn compositional information. To address this limitation, some approaches investigate how to augment the text captions or images in contrastive learning to enhance the ability of VLC~\cite{ours_teaching,aro}. Smith~\etal~\cite{seale2022construct} learn VLC concepts with additional supervised datasets in a continual learning setup. In contrast, we use 3D graphic engines to generate realistic synthetic videos with different compositions and generate corresponding text captions, which allows a VL model to learn compositionality and non-object words such as attributes, actions, relations, etc.

\noindent{\textbf{Learning from Synthetic Data.}} There has been a lot of work on learning from synthetic data  in image classification~\cite{TDW,peng2017visda,mishra2022task2sim}, semantic segmentation~\cite{ros2016synthia,richter2016playing}, human pose estimation~\cite{varol17_surreal,kim2022unified}, action recognition~\cite{varol21_surreact}, etc. Synthetic data is easy to generate and particularly useful for providing dense annotation such as semantic segmentation and depth estimation
since these are prohibitively expensive to annotate manually. 
Some of the work relies on graphics engines to generate realistic data. Mishra~\etal~\cite{mishra2022task2sim} propose a method to learn how to generate task-adaptive synthetic data with the 3D simulation engine. For human-related problems, parametric body models  (\eg, SMPL~\cite{loper2015smpl}) can be leveraged, along with motion assets~\cite{cmu_mocap}, to generate synthetic human videos for low-level body analysis tasks~\cite{varol17_surreal} or action recognition~\cite{DeSouzaCVPR2017,varol21_surreact}. Similar to our work, \cite{DeSouzaCVPR2017,varol21_surreact} seek to associate semantic labels to synthetic images, but different from symbolic action categories, our focus is to assign rich textual descriptions to our generated images.

Since synthetic data suffers from a domain gap such as textures, visual styles, or colors from real images, domain adaptation, and generalization have been proposed to address this issue. Adversarial learning~\cite{tzeng2017adversarial,ganin2016domain,shrivastava2017learning} can be used to generate real-like images or feature alignment between synthetic and real data. Additionally, stylization methods~\cite{zhou2021domain,zhao2022style}  are proposed as a style augmentation to make a model robust to diverse styles. In contrast, we manually randomize the visual content including different 3D objects, materials, and color attributes in graphics engines. Then we generate realistic synthetic videos from different domains with corresponding text captions. The generated data can be served as a hard negative augmentation and enhance the ability of VLC. 

\section{Method}
\label{sec:method}
We first present our synthetic data generation pipeline (Sec.~\ref{sec:syndata}), then describe how we leverage it for significant gains in \vlc{} understanding and compositional reasoning capabilities of strong pre-trained \vl{} models (Sec.~\ref{sec:training}). Our entire approach is illustrated in detail in Fig. \ref{fig:figCollidersTextures}.

\subsection{Synthetic Data Generation}\label{sec:syndata}
In this section, we outline the components and the pipeline of our approach used to generate the proposed \ourdatasetfull{} (\ourdataset{}) synthetic \vl{} dataset for improving \vlc{} understanding and compositional reasoning of \vl{} models. Our contributed dataset includes 767,738 image-text pairs, 598K sampled from 1,680 diverse synthetic videos, and the remaining 169K generated as individual static synthetic scenes. Example samples from \ourdataset{} are provided in Supplementary.

\noindent\textbf{3D physics-based simulation platform:} ThreeDWorld (TDW)~\cite{TDW}, which is built on top of Unity3D, is a multi-modal simulation platform that enables realistic physical interactions. 
TDW contains $2304$ objects, 585 unique materials subdivided in metal, cardboard, wood, ceramic, and glass, and over $30$ indoor and $9$ outdoor scenes (3D environments).  
For generating synthetic \vl{} data, we start with placing random objects in a scene following the workflow proposed by~\cite{9880308}. We also use their camera positions and configurations to place objects visible inside good empty room perspectives. We group the available 3D object models by assigning dimension-related labels to each object and use the ImageNet category labels available for each object model as its associated text for later caption synthesis.

\noindent\textbf{Camera Viewpoints}: To further augment the set of plausible object placements and relations, 
we simultaneously place $4$ to $12$ cameras around a specific point of an empty room, and randomly place $n \geq 1$ objects in the scene, allowing us to render images from different views of the same scene further strengthening the compositional aspects of the data as discussed in the introduction (Sec.~\ref{sec:intro}). For each scene (frame), TDW cameras are able to capture RGB images, the corresponding object instance and category semantic segmentation masks, and a depth map. We use these, as well as a range of sensor and physics data representing the state of the world returned by TDW's API, to enable dense annotations and supervision for each scene (frame) as part of our metadata generation process. We collect all of this information in our metadata and use it to estimate the position of the objects in the scene instead of relying on the 3D coordinates of each object and the camera position. 

\noindent\textbf{Digital humans}: As we focus on compositionality aspects of images and text pairs, having people in our images is important. However, people models (especially animatable ones) are usually not present in common collections of 3D assets. Existing large-scale synthetic datasets often focus on realistically placing objects in a scene, but typically humans and animals are not included. 
We first inspected what libraries were available for realistic human synthesis. PeopleSansPeople~\cite{ebadi2021peoplesanspeople}, a library with 28 human 3D models and 39 unique actions, allows only random human placement, not allowing for humanoid customization or integration of human-object interactions.
We leverage TDW support for Skinned Multi-Person Linear Model~\cite{SMPL} (SMPL) humanoids. SMPL is a parametric body model that enables a realistic representation of the shape and pose of arbitrary (non-clothed) 3D human bodies with diverse genders and shapes. SMPL models can be easily animated using motion capture data. The pose of the SMPL model is defined by a set of joint angles that determine the position of the corresponding body parts in the 3D space. The extended SMPL-X~\cite{SMPL-X} additionally allows controlling hand articulation and face expressions. Given the available library asset in TDW that enables placing these SMPLs in a scene, we create a stand-alone module to automatically incorporate arbitrary custom animations and $514$ unique human textures from the SURREAL~\cite{varol17_surreal} and Multi-Garment~\cite{bhatnagar2019mgn} datasets for clothing the synthetic human models for further enhancing the diversity and compositional features of our data.

\begin{figure}[h]
    \centering
    \includegraphics[width=\linewidth]{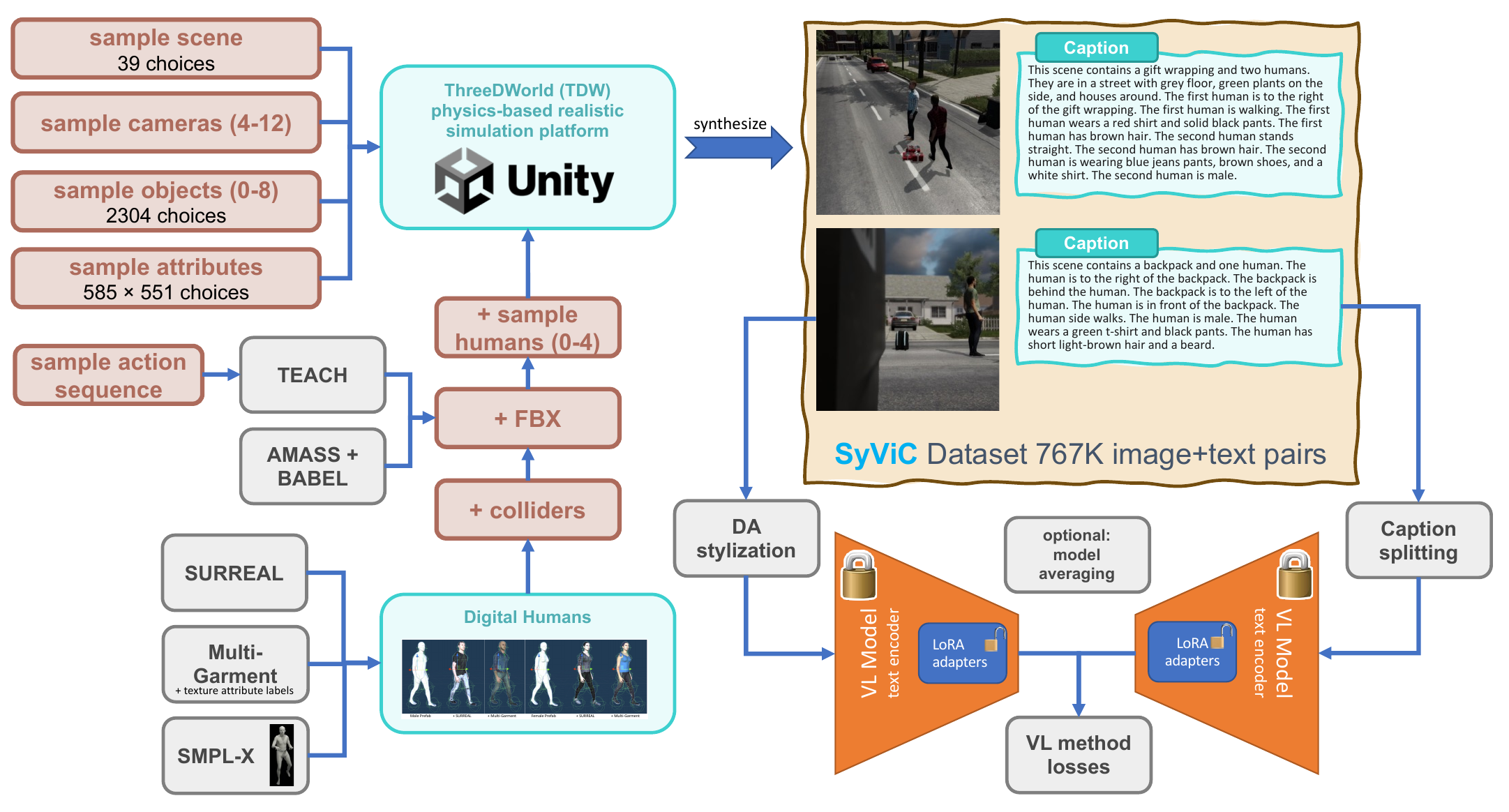} %{Images/figClothingSamples.pdf}
    \caption{Summarizing the entire flow of the proposed approach including components and choices of \ourdataset{} data synthesis pipeline \textbf{(left)} and proposed effective finetuning technique \textbf{(right)} attaining significant gains detailed below. } 
    \label{fig:figCollidersTextures}
    \vspace{-0.25cm}
\end{figure}

\noindent\textbf{Human motion synthesis and handling interactions:} In Unity, SMPLs have skeletons and can be driven by motion-capture animations but they don't have mass or colliders, this means that they are not physics assets since they can walk through other objects without interacting with them. 
To solve this issue, we add colliders to each body part of the SMPL model and create three asset templates (i.e., male, female, neutral) that contain all mesh configurations. Figure~\ref{fig:figCollidersTextures} shows some examples of our rigged asset templates with colliders. 
All other existing 3D models in TDW have colliders and track collisions at runtime through TDW physics-based simulation. Therefore, our collider-enhanced SMPLs are pulled downward by gravity, as well as simulate interaction by reacting naturally to collisions with other objects in the scene during motion simulation. 
For human actions, we first synthesize a diverse set of human actions from random language descriptions using TEACH~\cite{Athanasiou2022TEACHTA}, a Transformer-based model that generates a continuous sequence of SMPL motions from a sequence of action text labels. TEACH was trained on AMASS~\cite{AMASS}, a large-scale motion-capture (mocap) collection, and BABEL~\cite{BABEL}, a dataset that provides textual descriptions for AMASS, including per-frame unique actions annotations. Second, we extend our set of SMPL motions by directly sampling unique human motions from BABEL and AMASS. We export the corresponding mocaps to FBX files, and extract the animations in Unity, enabling them as asset bundles for use with TDW. FBX is a common format that facilitates data exchange between different 3D simulation platforms. 

\noindent\textbf{Domain randomization:} One of the key qualities of the generated synthetic data, is its ability to highlight the importance of \vlc{}s and compositional properties of the scene (e.g., objects attributes, relations, and more) in the contrastive learning objectives guiding the \vl{} model finetuning. As opposed to methods based on text augmentation~\cite{ours_teaching,aro} that can only enhance those on the text part of the \vl{}  image-text pairs, in \ourdataset{} construction we can easily manipulate also the visual content. We randomly place $1$ to $8$ 3D object models in the scene, randomizing their material and color attributes ($2304 \times 585 \times 551$ choices for each placed object). We randomly place $0$ to $4$ human avatars in the scene, randomizing their gender and clothing. We set camera poses as explained above, keeping scene ID shared for all the cameras of the same scene, and randomly sample $4$ viewpoints of each scene. We use human motion assets (as explained above) and randomly sample a motion sequence for each human avatar out of $1898$ imported or generated mocaps. Finally, we sample an average of $1500$ frames from each resulting synthetic video sequence generating image-text pairs describing a scene with the same objects and attributes, but in different arrangements and different corresponding captions thus enhancing the importance of the compositional aspects of the scene in the contrastive loss. We explore the importance of different simulation aspects in our ablation studies in Sec.~\ref{sec:ablations}.

\noindent\textbf{Metadata-driven caption text synthesis:} 
In addition to RGB frames, we obtain a large collection of rich metadata from the simulation platform, containing information on objects, humanoids, and the scene setting. For each frame, the metadata includes: (i) The world coordinates of each object and humanoid in the scene, including the camera position and viewing direction. (ii) The physical attributes of each object and humanoid in the scene (object physical attributes include color, size, and material; human attributes include the per-frame action label that changes over time and clothing description). (iii) Rendered depth images, instance segmentation masks, and category segmentation masks.
Using the metadata, we compute the positional relations between each pair of objects and/or humans by comparing the pixels covered by their segmentation masks as well as their camera coordinates. Then, we use a simple grammar that deterministically maps positional relationships, object attributes, human attributes and action descriptions, and scene descriptions to a well-formed caption. More details on the grammar are provided in Supplementary.

\subsection{Finetuning large-scale pre-trained \vl{} models using synthetic data} \label{sec:training}
In this section, we propose a methodology for effectively leveraging the \ourdataset{} synthetic \vl{} data produced as explained in Sec.~\ref{sec:syndata}. We will use the following notation. Let $(T,I)$ be the text~\&~image pair admitted by a \vl{} model. The model (e.g., CLIP~\cite{clip},  CyCLIP~\cite{cyclip})
components are denoted as: (i) image encoder $e_I = \mathcal{E}_I(I)$; (ii) text encoder $e_T = \mathcal{E}_T(T)$.
In this notation, 
the text-to-image similarity score is computed as:
\begin{equation}
    \mathcal{S}(T,I) = cos(\mathcal{E}_T(T), \mathcal{E}_I(I))) = cos(e_T,e_I),
\end{equation}
where $cos$ is the cosine similarity (inner product of normalized vectors).
We next describe in detail the components of our finetuning strategy. Their merit and tradeoffs are thoroughly investigated in Sec.~\ref{sec:ablations}, arriving at the conclusion that parameter efficient finetuning + domain adaptive stylization + proposed caption splitting technique are the most effective combination. We also confirm in Sec.~\ref{sec:ablations}, that model averaging can provide expected trade-offs between \vlc{} understanding and compositional reasoning gains and maintaining zero-shot performance.

\noindent\textbf{Avoiding forgetting through parameter efficient fine-tuning:}
Inspired by~\cite{ours_teaching,seale2022construct}, we use LoRA~\cite{lora} for \vl{} fine-tuning with reduced forgetting of base model performance. 
We apply LoRA~\cite{lora} to adapt the encoders ($\mathcal{E}_T$, $\mathcal{E}_I$)
of a pre-trained \vl{} model 
by parameterizing the adapted weights $\mathcal{W}_k^*$ 
corresponding to the original model weights $\mathcal{W}_k$ for each layer $k$ as:
\begin{equation}
    \mathcal{W}_k^* = \mathcal{W}_k + \mathcal{A}_k \cdot \mathcal{B}_k
\end{equation}
where for $\mathcal{W}_k$ of size $m \times l$, $\mathcal{A}_k$ and $\mathcal{B}_k$ are rank-$r$ matrices of sizes $m \times r$ and $r \times l$ respectively. These low-rank residual adapters can be applied efficiently during training and collapsed at inference time resulting in zero cost in terms of inference speeds or parameter counts~\cite{lora}.
During finetuning
all the base model 
parameters $\forall k,\{\mathcal{W}_k\}$ 
are frozen
and only the LoRA adapters $\forall k,\{(\mathcal{A}_k,\mathcal{B}_k)\}$ are being learned. 
Keeping rank $r$ low, the number of extra parameters added by all the LoRA adapters is low, consequently leading to significantly reduced forgetting in terms of largely maintaining the zero-shot performance of the original \vl{} model.

\noindent\textbf{Further reducing forgetting via model averaging:}~\cite{wortsman2022robust} introduced an elegant technique to mitigate forgetting in finetuned models. All the parameters of the source model (before finetune) and the final model (after finetune) are averaged between the two models (typically with $\alpha=0.5$ weight). We evaluate the effect of this on \ourdataset{} finetuned models in our ablation Sec.~\ref{sec:ablations}.

\noindent\textbf{Domain adaption using style transfer:} In addition, to mitigate the domain gap introduced by the use of synthetic data, we experiment with two style transfer techniques that align the content and feature statistics of the input frames with randomly-selected real-life images. A pre-trained Adaptive Instance Normalization (AdaIN)~\cite{adain} enabled encoder-decoder model was used to align the channel-wise statistics of each synthetic frame with a randomly-sampled image from the Human Motion Database (HMDB51)~\cite{hmdb51} dataset thus generating a stylized synthetic image. We use AdaIN with an interpolation factor $\alpha=0.5$. In addition, in order to preserve the color information in the synthetic frames, we first match the color distribution of the sampled style image to that of the synthetic frame~\cite{gatys_controlling}. 
We additionally experimented with MixStyle~\cite{mixstyle} (using ImageNet as a source of real style images) as an extension of the DA stylization pipeline without observing significant gains over AdaIN.

\noindent\textbf{Handling arbitrary caption length with caption splitting:} The captions generated for SyViC are comprehensive: they contain descriptions of every object and/or humanoid visible in the frame as well as the pairwise positional relationship between objects. Intuitively, including these more elaborate (dense) descriptions in our captions gives a clear advantage in terms of promoting \vlc{} understanding and compositionality following the finetuning of a \vl{} model on \ourdataset{}. 
Hence, captions need to be sufficiently long texts that cannot be fully processed by common \vl{} models (e.g. CLIP) text encoders ($\mathcal{E}_T$) during training, as those are caped by relatively short max sequence context length (e.g. 77 for CLIP). Therefore, inspired by CLIP multi-caption strategy for inference \cite{clip}, during training, we handle arbitrary caption lengths by splitting a given caption into sub-captions that can each be encoded separately and averaging the text features obtained from each sub-caption. In particular, the features of a caption of arbitrary length text $T$ is:
\begin{equation}
    \mathcal{E}_T(T) = \frac{1}{n} \sum_i^n \mathcal{E}_T(T_i)
\end{equation}
where $T_i$ is a sub-caption comprised of one or more sentences that fit into the text encoder max context size.

\noindent\textbf{Losses:} We employ the original models (e.g. CLIP~\cite{clip} and CyCLIP~\cite{cyclip})
contrastive and other losses when training on \ourdataset{} with the aforementioned architectural and training protocol changes as explained above.

\begin{table*}[]
\scriptsize
    \centering
    % \begin{small}
    \begin{tabular}{l|ll|l|llll|l|l}

            \toprule
            &\multicolumn{3}{c|}{VL Checklist } & \multicolumn{5}{c|}{ARO } & Zero-Short\\
            & Relation & Attribute & \textbf{Average} & VG-Rel. & VG-Att. & Flickr30k & COCO & \textbf{Average} & (21 tasks)\\
            \midrule            

            CLIP & 63.57 & 67.51 & 65.54 & 58.84 & 63.19 & 47.20 & 59.46 & 57.17 & 56.07\\
            CyCLIP & 61.15 & 66.96 & 64.06 & 59.12 & 65.41 & 20.82 & 29.54 & 43.72 & 55.99\\
            \midrule
            \ours{}CLIP & 69.39\gcol{+5.82} 	& 70.37\gcol{+2.86} 	& 69.88\gcol{+4.34} 	& 71.40\gcol{+12.56}	&	66.94\gcol{+3.75}	&	59.06\gcol{+11.86}	&	70.96\gcol{+11.5} & 67.09\gcol{+9.9} & 55.27\rcol{-0.8} \\
            \ours{}CyCLIP & 65.73\gcol{+4.58} & 68.06\gcol{+1.1} & 66.89\gcol{+2.83} & 69.02\gcol{+9.9} & 63.65\rcol{-1.76} & 49.17\gcol{+28.35} & 59.36\gcol{+29.82} & 60.30\gcol{+16.58} & 55.40 \rcol{-0.6}\\
            \bottomrule
    \end{tabular}
    
    % \end{small}
    \vspace{0.15in}
    \caption{Performance of \ours{}$<$\textit{model}$>$s -- finetuned on \ourdataset{} using our proposed recipe, measured on \vlchecklist{}~\cite{vlc} and \ARO{}~\cite{aro}. Gains and losses are highlighted in \green{green} and \red{red} respectively.}
    \label{tab:main_res_vl1}
    % \tabvspace
\end{table*}

\begin{table}[]
\scriptsize
    \centering
    % \begin{small}
    \begin{tabular}{l|lll|lll}

            \toprule
            & \multicolumn{3}{c|}{Winoground}& \multicolumn{3}{c}{Winoground$^\dag$}\\
            & Text & Image & \textbf{Group} & Text & Image & \textbf{Group}\\
            \midrule

            CLIP &  31.25 & 10.50 & 8.00 & 31.58 &	10.53 & 8.19\\
            \midrule
            \ours{}CLIP & 30.00 & 11.50 & 9.50\gcol{+1.50} & 29.82 & 12.28 & 9.94\gcol{+1.75} \\
            \bottomrule
    \end{tabular}
    
    % \end{small}
    \vspace{0.15in}
    \caption{\winoground{}~\cite{winoground} performance of \ours{}CLIP -- finetuned on \ourdataset{}.
    The \ours{}CyCLIP results on \winoground{} are provided in the Supplementary.
    $^\dag$ `clean' (no-tag) subset of valid \winoground{} samples from~\cite{why_is_winoground_hard}}
    \label{tab:main_res_vl2}
    \vspace{-0.2in}
\end{table}

\section{Experiments} \label{sec:exp}
\subsection{Implementation details}\label{sec:impl}
For CLIP, we use the original OpenAI CLIP implementation and checkpoints. We modify their codebase to include LoRA adapters (Sec. 3.2), and use rank 16 in all our experiments. For CyCLIP, we adapt the implementation used in~\cite{ours_teaching}~\footnote{Code and checkpoints kindly shared by the authors.}. For both CLIP and CyCLIP, we use a 5e-7 initial learning rate for finetuning and follow a cosine annealing learning rate schedule~\cite{cosine_annealing} using an Adam~\cite{adam} optimizer. For all experiments, we use ViT/32-B as the model architecture and fine-tune it for six epochs on one A100 GPU with a total batch size of $400$ image-caption pairs. In addition to the original CLIP data augmentation transforms, we apply heavy random augmentation policies including manipulations in image inversion, contrast, sharpness, equalization, posterization, colorization, brightness, and solarization.

\subsection{Datasets}\label{sec:datasets}
To test the effectiveness of our proposed \ourdataset{} synthetic dataset and the accompanying finetuning approach for improving \vl{} models' \vlc{} understanding and compositional reasoning capabilities we have evaluated on 3 benchmarks (\winoground{}~\cite{winoground}, \vlchecklist{}~\cite{vlc}, and \ARO{}~\cite{aro}) consisted of 7 datasets total.

\noindent\textbf{\vlchecklist{}~\cite{vlc}} -- is a large-scale dataset comprised of: Visual Genome~\cite{vg}, SWiG~\cite{swig}, VAW~\cite{vaw}, and HAKE~\cite{hake}. Each image of these datasets is associated with two captions, a positive and a negative. The positive caption corresponds to the image and is taken from the source dataset. The negative caption is made from the positive caption by changing one word, so the resulting sentence no longer corresponds to the image. Depending on the word that was changed, VL-Checklist evaluates 7 types of \vlc{} that can be divided into two main groups: (1) Attributes -- color, material, size, state, and action, and (2) Relations -- spatial or action relation between two objects and/or humans. In the following, we report average results for each of the main (Rel. and Attr.) groups on the combined \vlchecklist{} dataset. We also detail the individual improvements on all 7 \vlc{} types in Fig.~\ref{fig:radar} (left).

\noindent\textbf{\winoground{}~\cite{winoground}} -- is a small dataset that evaluates the ability of \vl{} models for compositional reasoning, specifically understanding the meaning of the sentence after changing the order of its words. The dataset has 400 samples, each comprised of two images and two texts. The texts have the same words in a different order, each text corresponding to one image in the sample. The \winoground{} metrics include (a) image score - percent of samples where the model picks the correct text for each image; (b) text score - percent of samples where the model picks the correct image for each text; (c) group score - percent of samples where both text and image score conditions are satisfied jointly. Recently,~\cite{why_is_winoground_hard} has analyzed \winoground{} for the source of its difficulty and found that only $171$ of its 400 samples are a valid subset. Other samples are not compositional, ambiguous, related to invisible details, have highly uncommon images or text, or require complex reasoning beyond compositionality. We report results on both the full \winoground{} and the `clean' 171 images subset from~\cite{why_is_winoground_hard}.

\noindent\textbf{\ARO{}~\cite{aro}} -- or the Attribution, Relation, and Order benchmark, is a large dataset designed to evaluate the ability of \vl{} models to understand four different types of skills. It consists of Visual Genome Attribution and Visual Genome Relation, which leverages the Visual Genome~\cite{vg} dataset along with the GQA~\cite{gqa} annotations to test the understanding of properties and relational understanding of objects in complex natural scenes. VG-Relation includes $48$ distinct relations with $23937$ test cases, and VG-Attribution includes $117$ unique attribute pairs with $28748$ test cases. It also leverages the COCO~\cite{coco} and Flickr30k~\cite{flickr30k} datasets to evaluate the model sensitivity to select the right caption after applying four different shuffling perturbations (e.g., exchanging nouns and adjectives, or by shuffling trigrams). These tests are performed on the $5000$ and the $1000$ images from the respective COCO and Flickr30k test splits.

\vspace{-.05em}
\subsection{Results}
\label{sec:results}
\begin{figure}[]
% \begin{minipage}[t]{0.45\textwidth}
%     (a)
    \includegraphics[width=0.45\linewidth]{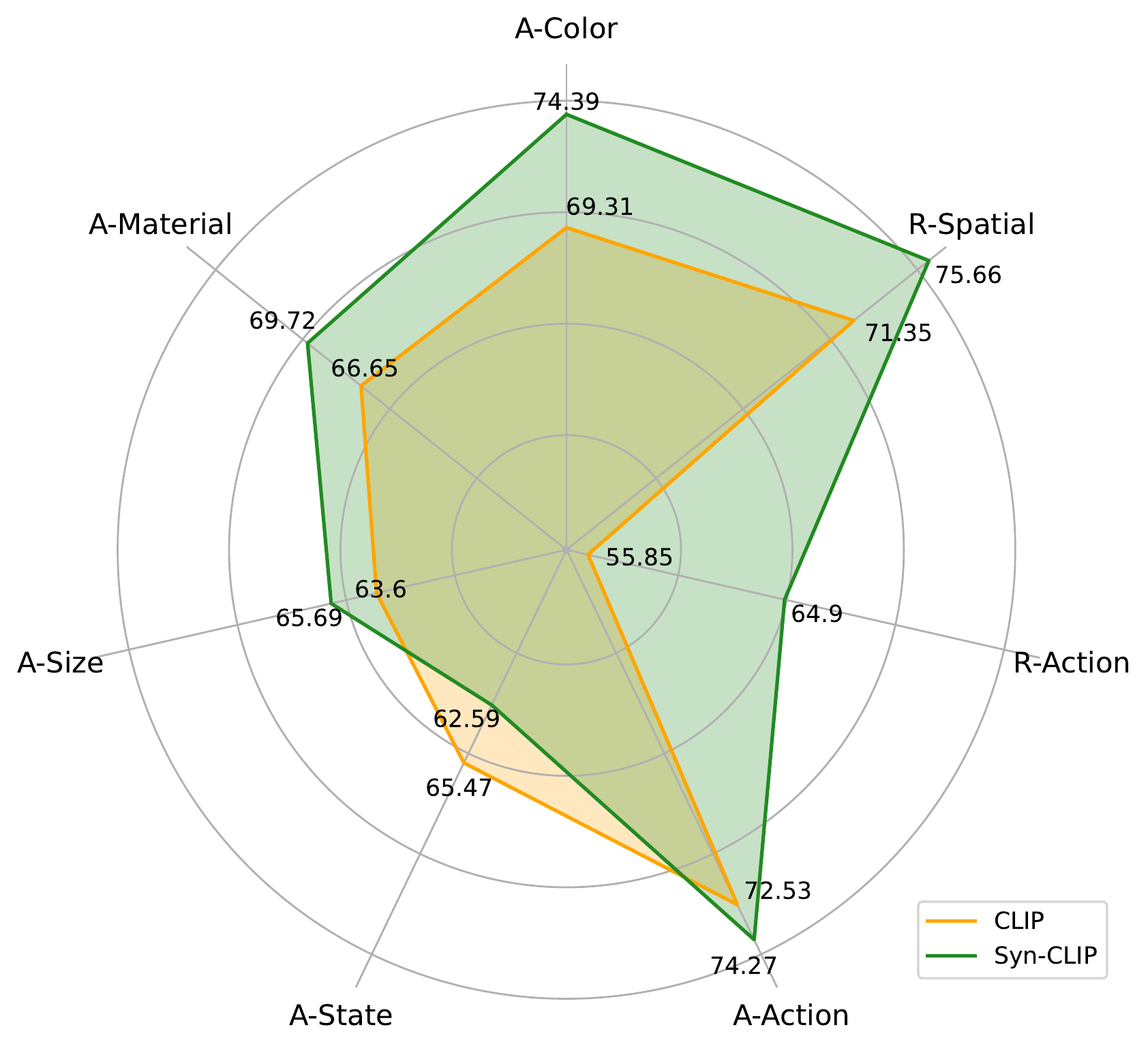}
% \end{minipage}%
\hfill % maximize the horizontal separation
% \begin{minipage}[b]{0.45\textwidth}
%  (b)
  \includegraphics[width=0.50\linewidth]{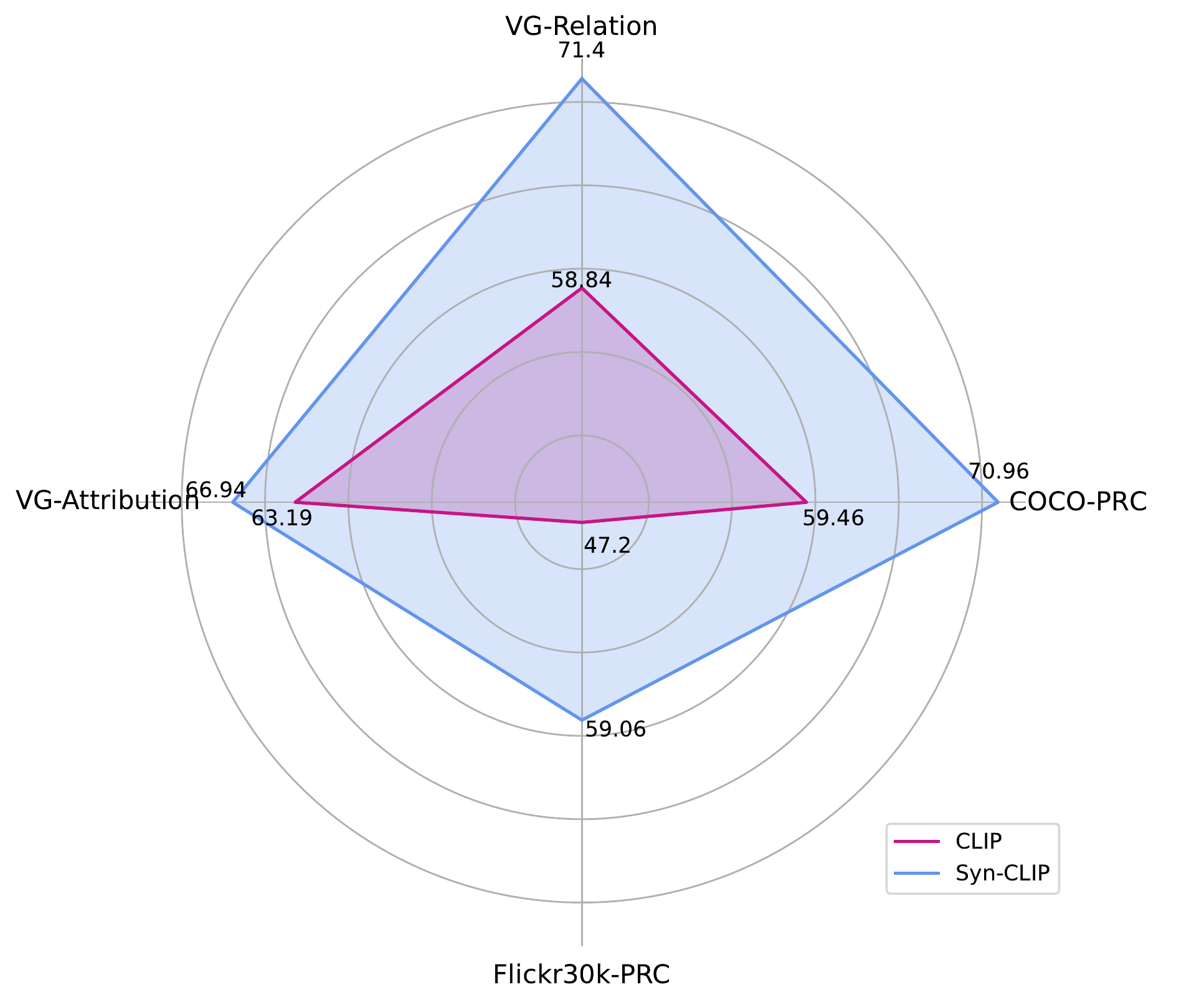}
% \end{minipage}%
\label{fig:abl-diversity}
    \caption{\textbf{(left)} detailed evaluation of \ours{}CLIP on the 7 separate \vlchecklist{}~\cite{vlc} metrics; \textbf{(right)} detailed evaluation of \ours{}CLIP on all the Compositional Tasks proposed in \ARO{}~\cite{aro}. } 
    \label{fig:radar}
    \vspace{-1em}
\end{figure}

The main results of finetuning CLIP~\cite{clip}, CyCLIP~\cite{cyclip} -- one of CLIP's most recent improvements
% , and BLIP~\cite{blip} 
are summarized in Tables~\ref{tab:main_res_vl1} and~\ref{tab:main_res_vl2}. All models were finetuned using our proposed approach and \ourdataset{} synthetic data to obtain their \ours{}$<$\textit{model}$>$ variants. Each model is compared to its respective source model pre-trained on large-scale real data before finetuning on \ourdataset{}. As we can observe, our \ourdataset{} synthetic data and the proposed finetuning recipe on this data demonstrate significant improvements over their source baselines. E.g. for CLIP obtaining $1.75\%$, $4.34\%$, and $9.9\%$ average absolute improvement in \winoground{} group score (most difficult average metric), \vlchecklist{} and \ARO{} respectively. In addition, we illustrate the individual \vlc{} metrics improvements obtained for CLIP in \vlchecklist{} and \ARO{} benchmarks in Fig.~\ref{fig:radar} showing up to $9.1\%$ and $12.6\%$ respective absolute improvements. This underlines the effectiveness and promise of our method and \ourdataset{} synthetic data towards improving \vlc{} understanding and compositional reasoning in \vl{} models. Importantly, as we can see from Table~\ref{tab:main_res_vl1}, these strong gains come at a very small (under 1\%) cost in the zero-shot performance of the respective \vl{} models measured using the standard Elevater~\cite{elevater} benchmark using 21 diverse zero-shot tasks.

\begin{table}[]
\scriptsize
    \centering
    % \begin{small}
    \begin{tabular}{cc|ll|l}
            \toprule
            objects with attr. & humans & \multicolumn{3}{c}{VL Checklist} \\
            randomization & & Relation & Attribute & Average \\
            \midrule            
            \multicolumn{2}{c|}{CLIP} & 63.57 & 67.51 & 65.54 \\ 
            \midrule
            \xmark\ & \cmark\ & 64.03	& 67.09	& 65.56 \\ 
            \cmark\ & \xmark\ & 65.00 & 68.15 & 65.95 \\ 
            \cmark\ & \cmark\ & \textbf{69.39} & \textbf{70.37} & \textbf{69.88} \\

            \bottomrule
    \end{tabular}
    % \end{small}
    \vspace{0.15in}
    \caption{Importance of human avatars, objects, and object attribute variations, evaluated on \vlchecklist{} and CLIP
    }
    \label{tab:abl-humans-objects}
    % \tabvspace
    \vspace{-2em}
\end{table}

\subsection{Ablations}\label{sec:ablations}
We extensively ablate our \ourdataset{} synthetic data and the proposed \vl{} models finetuning approach on this data according to the following points. We use the most popular CLIP model finetuned on our \ourdataset{} synthetic dataset evaluated on the largest of the benchmarks - the \vlchecklist{} to perform our ablations.

\noindent\textbf{\ourdataset{} - objects, humans, object attribute randomization} -- we evaluate the major components that comprise our \ourdataset{} synthetic data, namely the importance of the synthetic data to contain humans performing various motions and actions, the importance of having objects with randomized attributes (Sec.~\ref{sec:syndata}), and the final result of having all types of data combined. The results of this ablation are summarized in Tab.~\ref{tab:abl-humans-objects}. As expected, humans alone cannot teach the model the needed skills only improving relations \vlc{} by a small margin. Additionally, having only objects with randomized attributes improves attribute \vlc{}, yet only improves relations by $1.4\%$ which is also expected, as many of the relations involve human actions. The best result is observed on the combined dataset with all the components.

\begin{table}[]
\scriptsize
    \centering
    % \begin{small}
    \begin{tabular}{cc|ll|l}
            \toprule
            SURREAL & Multi-Garment & \multicolumn{3}{c}{VL Checklist} \\
            & & Relation & Attribute & Average \\
            \midrule            
            \multicolumn{2}{c|}{CLIP} & 63.57 & 67.51 & 65.54  \\
            \midrule
            \xmark\ & \xmark\ & 67.56 & 68.73 & 68.15 \\ 
            \cmark\ & \xmark\ & 67.56 	& 67.26 	& 67.41 \\
            \xmark\ & \cmark\ & \textbf{69.39} & \textbf{70.37} & \textbf{69.88} \\
            \bottomrule
    \end{tabular}
    % \end{small}
    \vspace{0.15in}
    \caption{Importance of human avatar clothing choices between SURREAL, Multi-Garment, and simple color textures (corresponding to none), evaluated on \vlchecklist{} and CLIP}
    \label{tab:abl-clothing}
    % \tabvspace
    \vspace{-0.1in}
\end{table}

\noindent\textbf{\ourdataset{} - human clothing} -- we evaluate the diversity of human clothing comparing 3 levels of diversity: (i) none - using only a uniform color for human models; (ii) basic - using less diverse texture maps from SURREAL~\cite{varol17_surreal}; and (iii) most diverse - using texture maps from Multi-Garment~\cite{bhatnagar2019mgn}, enriched with clothing colors, human age, and hair color annotations (manually done by us for the textures) which increase captions' expressivity. Results are presented in Table~\ref{tab:abl-clothing}. As expected, the most diverse human textures deliver the best result underlining the importance of this factor. Surprisingly, better human textures improve \vlchecklist{} Relations metric performance, likely due to the significantly better realism of the Multi-Garment textures. 

\noindent\textbf{\ourdataset{} - types of object attributes to randomize} -- Table~\ref{tab:abl-attributes-randomization} examines how randomizing different object attributes affects performance. Specifically, we evaluate the randomization of size, material, and color. Interestingly, we find that the best performance is achieved without color randomization. We suspect it is due to unnatural color-object combinations that arise under such randomization, which teach the model wrong beliefs on real objects' color distributions and go against true object-color associations existing in the \vl{} model following pre-training on the original \vl{} data.

\noindent\textbf{\ourdataset{} - types of captioning} -- we have investigated several variants of ways to obtain textual captions from \ourdataset{} metadata (Sec.~\ref{sec:syndata}). Results are summarized 
the Supplementary.
We compared our proposed metadata grammar-based approach to two cascade methods that paraphrase the captions resulting from the grammar using zero-shot LLM inference (in-context learning). The paraphrased caption is then appended to the original grammar-based caption and consumed through our caption-splitting module (as standalone, open LLM-based paraphrasing is not very high quality). As can be seen, currently paraphrasing has minimal effect, but we posit it will become an important tool as stronger LLMs will become openly available.

\noindent\textbf{\ourdataset{} - importance of physics and number of humans in the scene} -- we also looked into to which extent reliable physical simulation (made available in \ourdataset{} through TDW and Unity capabilities) and human-human positional and other relations are important for the observed \vl{} model improvements. In Table~\ref{tab:abl-physics-interactions} we evaluate the effects of removing the physics (resulting in humans or objects floating in space) or removing the multi-human scenes (thus preventing all human-human relations from appearing in the data). As expected, both reliable physics simulation and human-human relations (interactions of sorts) are mostly important to the gains in the Relations metric.

\begin{table}[]
\scriptsize
    \centering
    % \begin{small}
    \begin{tabular}{ccc|ll|l}
            \toprule
            color & size & material & \multicolumn{3}{c}{VL Checklist} \\
            & & & Relation & Attribute & Average  \\
            \midrule            
            \multicolumn{3}{c|}{CLIP} & 63.57 & 67.51 & 65.54  \\
            \midrule
            \cmark\ & \xmark\ & \xmark\ & 67.71 & 64.61 & 66.16  \\ 
            \xmark\ & \cmark\ & \xmark\ & 68.58 & 68.23 & 68.40  \\ 
            \xmark\ & \xmark\ & \cmark\ & 65.23 & 67.01 & 66.12  \\ 
            \cmark\ & \cmark\ & \cmark\ & 66.67 & 65.97 & 66.32 \\ 
            \xmark\ & \cmark\ & \cmark\ & \textbf{69.39} & \textbf{70.37} & \textbf{69.88} \\ 
            \bottomrule
    \end{tabular}
    % \end{small}
    \vspace{0.15in}
    \caption{Importance of different kinds of object attributes randomization, evaluated on \vlchecklist{} and CLIP}
    \label{tab:abl-attributes-randomization}
    % \tabvspace
    \vspace{-0.1in}
\end{table}

\noindent\textbf{\ourdataset{} - number of models and number of samples} -- for lack of space, this is explored in the Supplementary.

\begin{table}[b]
\scriptsize
    \centering
    % \begin{small}
    \begin{tabular}{cc|ll|l}
            \toprule
            physics & multi-human & \multicolumn{3}{c}{VL Checklist} \\ 
            & & Relation & Attribute & Average  \\
            \midrule            
             \multicolumn{2}{c|}{CLIP} & 63.57 & 67.51 & 65.54 \\ 
            \midrule
            \cmark\ & \xmark\ & 67.66 & 69.24 & 68.45 \\ 
            \xmark\ & \cmark\ & 65.91 & 69.01 & 67.46 \\ 
            \cmark\ & \cmark\ & 69.39 & 70.37 & 69.88 \\ 
            \bottomrule
    \end{tabular}
    % \end{small}
    \vspace{0.15in}
    \caption{Importance of physics simulation and multi-human relations in \ourdataset{}, evaluated on \vlchecklist{} and CLIP}
    \label{tab:abl-physics-interactions}
    % \tabvspace
\end{table}

\begin{table*}[]
\scriptsize
    \centering
    % \begin{small}
    \begin{tabular}{l|ccccc|ll|l|c|c}
            \toprule
            \# & LoRA & freezing & model & DA & Caption &\multicolumn{3}{c|}{VL Checklist} & ARO & ZS \\
            & & $\mathcal{E}_I$ & averaging & styl. & split emb. & Rel. & Attr. & Avg. & Avg. & (21 tasks) \\
            \midrule            
            & \multicolumn{5}{c|}{CLIP} & 63.57 & 67.51 & 65.54 & 57.17 & 56.07 \\
            \midrule
            1 & \xmark\ & \xmark\ & \xmark\ & \xmark\ & \xmark\ & 67.10 & 65.45 & 66.28 & 62.83 & 53.84 \\
            2 & \xmark\ & \xmark\ & \xmark\ & \xmark\ & \cmark\ & 67.76 & 67.51 & 67.64 & 59.27 & 53.87 \\
            3 & \cmark\ & \xmark\ & \xmark\ & \xmark\ & \cmark\ & \blue{69.32} & \blue{69.46} & \blue{69.39} & \blue{64.34} & 53.16 \\
            \midrule
            4 & \cmark\ & \xmark\ & \xmark\ & \cmark\ & \cmark\ & \textbf{69.39} & \textbf{70.37} & \textbf{69.88} & \textbf{67.09} & \blue{55.27} \\
            \midrule
            5 & \cmark\ & \cmark\ & \xmark\ & \cmark\ & \cmark\ & 63.54 & 68.20 & 65.87 & 60.29 & 54.54 \\
            6 & \cmark\ & \xmark\ & \cmark\ & \cmark\ & \cmark\ & 66.70 & \blue{69.62} & 68.16 & 63.76 & \textbf{55.72} \\
            7 & \cmark\ & \cmark\ & \xmark\ & \cmark\ & \xmark\ & 65.16 & 66.88 & 66.02 & 57.55 & 52.69 \\
            \bottomrule
    \end{tabular}
    % \end{small}
    \vspace{0.15in}
    \caption{Importance of the finetuning recipe components, evaluated on \vlchecklist{} and CLIP}
    \label{tab:abl-finetuning-recipe}
    \vspace{-0.15in}
    % \tabvspace
\end{table*}

\noindent\textbf{\ourdataset{} - finetuning recipe components} -- in Table~\ref{tab:abl-finetuning-recipe} we extensively evaluate the different components of the proposed finetuning approach on \ourdataset{} that leads to significant improvements on \vlchecklist{}, \ARO, and \winoground{} \vlc{} and compositional reasoning metrics. We start with vanilla CLIP finetuning on \ourdataset{} (row \#1), already showing some improvement in \vlchecklist{} relations metrics, and on the \ARO{} benchmark, yet losing to base CLIP on \vlchecklist{} attributes metrics. Adding our caption splitting module (Sec.~\ref{sec:training}) allows handling long (arbitrary size) texts outputted by our metadata-driven grammar and consequently utilizes all the caption information re-gaining the attributes performance (row \#2). Adding parameter-efficient finetuning (LoRA, Sec.~\ref{sec:training}) regularizes finetuning by forcing smaller (low-rank, low-parameters) updates of the large-scale pre-trained CLIP model, consequently somewhat handling the expected domain gap between the synthetic data of \ourdataset{} and the downstream evaluation (real data) tasks. Notably, LoRA does not add any additional parameters to the model, all LoRA adapters are collapsed into the model weights after finetuning. Consequently, we observed significant improvements from adding LoRA in all metrics (row \#3) with only minor degradation (0.7\%) in ZS evaluation. With adding domain stylization (Sec.~\ref{sec:training}) we observe the best results in all \vlc{} and compositional reasoning metrics improving \ARO{} by 2.8\% and keeping (even slightly improving) the advantages on \vlchecklist{}. Next, we investigate the variations of our best approach (LoRA + domain stylization + caption splitting). First, we investigate a strategy inspired by the LiT~\cite{lit} approach (row \#5). Freezing the visual encoder $\mathcal{E}_I$ as expected provides a (small) boost in ZS performance, but the reduced plasticity of the model comes at the price of observing smaller (only 3.1\%) improvements on \ARO{} and almost no improvements on the \vlchecklist{}. This leads us to conclude, that freezing the visual encoder is not a good strategy for \ourdataset{} finetuning. Next, we check the model averaging strategy (Sec.~\ref{sec:training}) inspired by~\cite{wortsman2022robust} (row \#6). This does a better job of mitigating ZS forgetting, while at the same time keeping more of the gains on \vlchecklist{} and \ARO{}. We conclude that model averaging is a good strategy to complement \ourdataset{} finetuning, allowing a soft trade-off between mitigating ZS forgetting and \vlc{} and compositionality metrics gains. Finally, we again explore the importance of caption splitting for the best finetuning configuration of \ourdataset{} (row \#7) and re-confirm its significance as performance drops without it.

\section{Summary \& Conclusions}
\label{sec:conclusion}
Large vision and language models have dictated the status quo in computer vision and multimodal perception, achieving state-of-the-art results in a number of challenging benchmarks. However, existing models struggle with compositional reasoning and understanding concepts beyond object nouns, such as attributes and relationships. Our work has investigated, for the first time, whether synthetic data can be leveraged to mitigate these shortcomings. We proposed a data generation pipeline, used to create a million-scale dataset of synthetic images and accompanying captions, and an effective fine-tuning strategy with comprehensive analysis to enhance the compositional and concept understanding capabilities of multimodal models, without compromising their zero-shot classification performance.

\vspace{0.02in}
\noindent\textbf{Limitations.}  While we have achieved quite promising results in three different benchmarks, our work has limitations. As an example, our graphics simulator has a simplified model of lighting, sensor noise, and reflectance functions compared to the real world, which may impact robustness to color constancy. We believe more advanced domain adaptation and rendering techniques are likely needed to further improve our results. We also think a more detailed study of the scaling laws for synthetic data is a great research direction to fully unlock the potential of our work.

\vspace{1mm}
{
\small
\noindent\textbf{Acknowledgements.}
This material is based upon work supported by the Defense Advanced Research Projects Agency (DARPA) under Contract No. FA8750-19-C-1001. Any opinions, findings and conclusions or recommendations expressed in this material are those of the author(s) and do not necessarily reflect the views of the Defense Advanced Research Projects Agency (DARPA). This project was also partially supported by NSF Award IIS-2221943, ANR CorVis ANR-21-CE23-0003-01, and the MIT-IBM Watson AI Lab. We thank Jeremy Schwartz and Esther Alter for their helpful discussions and assistance with the TDW package.
\par
}

{\small
\bibliographystyle{ieee_fullname}
\bibliography{references}
}

\newpage
\appendix

\section*{Appendix}
In this supplementary material, we share our code and provide additional insights and experimental results that were not included in the main paper due to space constraints. In Section~\ref{sec:code} we describe the implementation code for generating \ourdataset{} and the code for the proposed finetuning approach on \ourdataset{}.
In Section~\ref{sec:more_models_on_vlc}, we analyze the performance of recently open-sourced \vl{} models on VL-Checklist \cite{vlc} and show they have low performance, demonstrating the need for our improvements. 
In Section~\ref{sec:sm:wino}, we provide additional results on \winoground{} \cite{winoground} using CyCLIP \cite{cyclip} (excluded from the main text for space purposes). 
Section~\ref{sec:blip_extra} demonstrates how we can improve BLIP \cite{blip} using \ourdataset{}. 
In Section~\ref{sec:with_neg}, we combine our contributions with those of \textit{concurrent} work of \cite{ours_teaching} and demonstrate that our approach for improving \vlc{} using synthetic data is \textit{orthogonal / complementary} to the text-augmentation based methods. 
Section~\ref{sec:sm:grammar} provides more dataset details, describing how the metadata from each synthesized scene is used to generate a caption for each image in \ourdataset{}. 
In Section~\ref{sec:sm:llms}, we explore combinations of our metadata-driven grammar-based caption generation with paraphrasing using openly available large language models. Section \ref{sec:sm:diversity} provides "\ourdataset{} - number of models and number of samples" ablation excluded from the main paper due to lack of space.
Finally, in Section~\ref{sec:sm:qual} we provide some randomly sampled examples from \ourdataset{}. 

\section{Code}\label{sec:code}
Our code for both \ourdataset{} data synthesis and the proposed finetuning approach is included in our project page: \href{https://synthetic-vic.github.io/}{https://synthetic-vic.github.io/} 

\section{Expanding VL-Checklist \cite{vlc} analysis to most recent \vl{} models}\label{sec:more_models_on_vlc}
As promised in the footnote in the introduction we have evaluated the very recently released open-source \vl{} models, namely: METER (CVPR 22) \cite{meter}, X-VLM (ICML 22) \cite{xvlm}, and VLMO (NeurIPS 22) \cite{vlmo} 
on the most extensive \vlc{} understanding benchmark of \vlchecklist{} \cite{vlc} observing average performance of 56.8\%, 58.9\%, and 54.6\%
respectively. As noted in the introduction, this relatively low \vlc{} understanding performance (below CLIP \cite{clip}) of the newest (open) \vl{} models illustrates once again the very much needed improvement in this aspect. Consequently, it also underlines the importance of \ourdataset{} and the proposed finetuning approach for administering some of this improvement and highlighting the future potential of our approach and synthetic data in general for the \vl{} modeling. We additionally explore the very recent BLIP \cite{blip} model and how it could be improved using \ourdataset{} and our approach in Section~\ref{sec:blip_extra}.

\section{Winoground Results for CyCLIP \cite{cyclip}}\label{sec:sm:wino}
As promised in the main paper (lines 555-556), we include the \winoground{} \cite{winoground} results of CyCLIP \cite{cyclip} not included in the main paper for lack of space. The results are included in Table \ref{tab:winoground_cyclip} and, compared to the CyCLIP baseline, demonstrate stable improvements of up to $1.17\%$ group score for \ours{}CyCLIP finetuned on \ourdataset{} using our proposed approach.
\begin{table*}[]
\scriptsize
    \centering
    \begin{small}
    \begin{tabular}{l|ll|l|ll|l}

            \toprule
            & \multicolumn{3}{c|}{Winoground}& \multicolumn{3}{c}{Winoground$^\dag$}\\
            & Text & Image & \textbf{Group} & Text & Image & \textbf{Group}\\
            \midrule

            CyCLIP & 28.50 & 9.50 & 7.25 & 32.16 & 11.11 & 8.19 \\
            \midrule
            \ours{}CyCLIP & 30.00\gcol{1.5} & 10.75\gcol{1.25} & 8.25\gcol{1.0} & 30.99\rcol{-1.17} & 12.87\gcol{1.76} & 9.36\gcol{1.17} \\
            \bottomrule
    \end{tabular}
    
    \end{small}
    \vspace{0.15in}
    \caption{\winoground{} \cite{winoground} performance of \ours{}CyCLIP -- finetuned on \ourdataset{}. 
    $^\dag$ `clean' (no-tag) subset of valid \winoground{} samples from \cite{why_is_winoground_hard}
    }
    \label{tab:winoground_cyclip}
\end{table*}
\section{Improving BLIP \cite{blip} using \ourdataset{}}\label{sec:blip_extra}
BLIP is a recently released \vl{} model achieving better out-of-the-box performance on \vlchecklist{} \cite{vlc} and \winoground{} \cite{winoground} compared to CLIP \cite{clip}. In Table \ref{tab:blip_extra} we show how using our proposed \ourdataset{} dataset and the finetuning approach applied to BLIP, significant additional performance gains ($1.48\%$ on \vlchecklist{} and up to $4.67\%$ on \winoground{} group score) can be achieved. 
BLIP is designed and optimized for \vl{} understanding and generation \cite{blip}, and has a relatively low zero-shot out-of-the-box performance compared to CLIP (e.g., we observed an over $7\%$ drop in zero-shot comparing baseline BLIP to CLIP and similar drop for \ours{}BLIP compared to CLIP). 
In more detail,
we employ the retrieval flow of BLIP starting from ViT/B and CapFilt-L base model and use it as the BLIP baseline in Table \ref{tab:blip_extra}. We finetune BLIP on \ourdataset{} following the complete proposed recipe detailed in 
% Section~3.2 
Section~\ref{sec:training}
% \if\sepappendix1{Section~3.2}
% \else{Section~\ref{sec:training}}
of the main paper. We add rank-16 LoRa adapters to both BLIP encoders and the decoder (cross-attention layers in the text encoder). 
We fine-tune for two epochs with a learning rate of 5e-6 using an Adam optimizer with a weight decay factor of $0.05$ \cite{adamw}. 

\begin{table*}[t!]
\scriptsize
    \centering
    \begin{tabular}{l|ll|l|ll|l|ll|l}
            \toprule
            &\multicolumn{3}{c|}{VL Checklist } & \multicolumn{3}{c|}{Winoground} & \multicolumn{3}{c}{Winoground$^\dag$} 
            \\
            & Relation & Attribute & \textbf{Average} & Text & Image & \textbf{Group} & Text & Image 
            \\
            \midrule  
            BLIP & 68.45 & 73.11 & 70.78 & 38.00 & 18.25 & 14.50 & 43.86 & 25.15 & 21.06 
            \\
            \midrule
            \ours{}BLIP & 70.18\gcol{+1.73} & 75.34\gcol{+2.23} & 72.76\gcol{+1.48} & 43.25\gcol{+5.25} & 19.75\gcol{+1.5} & 16.75\gcol{+2.25} & 52.63\gcol{+8.77} & 29.82\gcol{+4.67} & 25.73\gcol{+4.67} 
            \\
            \bottomrule
    \end{tabular}
    
    \vspace{0.15in}
    \caption{Performance of \ours{}BLIP -- finetuned on \ourdataset{} and evaluated on VL-Checklist and Winoground. $^\dag$ `clean' (no-tag) subset of valid \winoground{} samples from \cite{why_is_winoground_hard}. Gains and losses are highlighted in \green{green} and \red{red} respectively.}
    \label{tab:blip_extra}
\end{table*}

\section{Exploring a combination with text augmentation methods}\label{sec:with_neg}
As discussed in lines 108-112 in the Introduction of the main paper, \textit{concurrent} works \cite{ours_teaching,aro} propose an \textit{orthogonal} approach of improving \vlc{} understanding performance via using text augmentation while training on additional real \vl{} paired image+text data. These works use language tools to teach a model the importance of non-noun words by manipulating them (replacing words with incorrect alternatives in the text captions of real image+text pairs) and adding the resulting texts to the same batch. In order to show that our proposed approach of improving the \vlc{} understanding performance of \vl{} models using targeted demonstration on \textit{both text and image side} via generating synthetic data (our \ourdataset{} dataset) is truly orthogonal and complementary to the text augmentation methods, we have conducted the following experiment whose results are summarized in Table \ref{tab:comb_with_neg}. Specifically, we used \cite{ours_teaching} code, kindly shared to us by the authors, to combine our \ourdataset{} finetuning (as described in 
% Section~3.2 
Section~\ref{sec:training}
% \if\sepappendix1{Section~3.2}
% \else{Section~\ref{sec:training}}
of the main paper) with the LAION \cite{laion} experiment of \cite{ours_teaching} using their text augmentation method both for the real data captions as well as for our \ourdataset{} synthetic data captions. More specifically, we finetune (using the method described in 
% Section~3.2 
Section~\ref{sec:training}
% \if\sepappendix1{Section~3.2}
% \else{Section~\ref{sec:training}}
in the main paper, also including \cite{ours_teaching} negative text augmentations and their additional losses for the negatives as detailed in \cite{ours_teaching}) on combined batches containing both LAION \cite{laion} text+image pairs and \ourdataset{} text+image pairs.  The base model is CLIP \cite{clip} both for \cite{ours_teaching} and \ours{}\cite{ours_teaching}. As we can see in \ref{tab:comb_with_neg}, \ours{}\cite{ours_teaching} significantly (up to $12.26\%$ on Relations and $8.46\%$ on average) improves the base \cite{ours_teaching} performance on \vlchecklist{} (trained on the same LAION data without \ourdataset{}) and is roughly matching \cite{ours_teaching} performance on the \ARO{} and zero-shot evaluations.

\begin{table*}[t!]
\scriptsize
    \centering
    \begin{tabular}{l|ll|l|llll|l|l}
            \toprule
            &\multicolumn{3}{c|}{VL Checklist } & \multicolumn{5}{c|}{ARO } & Zero-Short\\
            & Relation & Attribute & \textbf{Average} & VG-Rel. & VG-Att. & Flickr30k & COCO & \textbf{Average} & (21 tasks)\\
            \midrule            
            CLIP & 63.57 & 67.51 & 65.54 & 58.84 & 63.19 & 47.20 & 59.46 & 57.17 & 56.07\\
            \cite{ours_teaching} & 66.05 & 69.64 & 67.85 & 80.64 & 72.81 & 92.82 & 87.67 & 83.48 & 56.71 \\
            \midrule
            \ours{}CLIP & 69.39\gcol{+5.82} 	& 70.37\gcol{+2.86} 	& 69.88\gcol{+4.34} 	& 71.40\gcol{+12.56}	&	66.94\gcol{+3.75}	&	59.06\gcol{+11.86}	&	70.96\gcol{+11.5} & 67.09\gcol{+9.9} & 55.27\rcol{-0.8} \\
            \ours{}\cite{ours_teaching}
            & 78.31\gcol{+12.26} & 74.31\gcol{+7.67} & 76.31\gcol{+8.46} & 80.79\gcol{+0.15} & 72.37\rcol{-0.44} & 92.44\rcol{-0.38} & 87.19\rcol{-0.48} & 83.20\rcol{-0.28} & 54.57\rcol{-2.14} \\
            \bottomrule
    \end{tabular}
    
    \vspace{0.15in}
    \caption{
    Demonstrating that text augmentation on real paired \vl{} data training is orthogonal/complementary to our approach. Comparing \cite{ours_teaching} performance finetuned on LAION with \ours{}\cite{ours_teaching} performance finetuned on LAION + \ourdataset{} using a combination of our approach in 
    % Section~3.2 
    Section~\ref{sec:training}
    % \if\sepappendix1{Section~3.2}
    % \else{Section~\ref{sec:training}}
    of the main paper with \cite{ours_teaching}'s negatives text augmentations and additional negative losses (added from \cite{ours_teaching} original code kindly shared with us by the authors of \cite{ours_teaching}). The base model is CLIP \cite{clip} both for \cite{ours_teaching} and \ours{}\cite{ours_teaching}.
    Results are evaluated on \vlchecklist{} \cite{vlc} and \ARO{} \cite{aro}. 
    Gains and losses are highlighted in \green{green} and \red{red} respectively. \ours{}\cite{ours_teaching} is comparable on \ARO{} (with $0.28\%$ difference) and significantly improving on \vlchecklist{} (with $8.46\%$ average improvement), while having only a small decrease on the zero-shot evaluation, only $2.14\%$ w.r.t. \cite{ours_teaching} and even smaller $1.5\%$ compared to base CLIP model.
    }
    \label{tab:comb_with_neg}
    % \tabvspace
\end{table*}

\section{Metadata-driven caption text synthesis, more details}\label{sec:sm:grammar}
This section describes how the metadata from each synthesized scene is used to generate a caption for each image in \ourdataset{}. We outline a rule-based mechanism to deterministically generate dense captions given:
\begin{enumerate}
    \item List of the objects present in the scene, each with its corresponding name and world coordinates.
    \item List of humanoids present in the scene, each with its world coordinates, clothing identifier, and a textual description of the action it performs.
    \item Segmented image that has a label for each pixel corresponding to the object or humanoid it belongs to.
    \item Scene identifier that maps to a textual description of the scene.
\end{enumerate}

We annotate a list of 115 clothing textures from the Multi-Garment ~\cite{bhatnagar2019mgn} and SURREAL ~\cite{varol17_surreal}. Clothing annotations include a list of textual descriptions such as the colors of the clothes, the hair/beard style, as well as any features that stand out such as logos, tattoos, and accessories. Additionally, we use the original scene descriptions provided by ThreeDWorld's scene library.

To generate the description of the objects in the image, we use the (3D) world positions of the objects to create positional relations between them. For objects that are horizontally aligned, we generate a description of which object is to the left or right of the other by comparing their corresponding pixels in the segmentation image. Furthermore, we generate a description of which object is in front of the other by translating the world coordinates into camera coordinates and comparing their z-coordinates. Unique identifiers for names are used to as placeholders to obtain those relationships, and object names are filled in once all positional relations are established, using indefinite articles when necessary. This process is applied to every pair of objects present in the scene.

Furthermore, we generate descriptions of humans while referring to them using ordinal numbers. In particular, for each human present in the scene, we retrieve its action description and place it in a sentence (e.g. "The \{first\} person \{walks forward\}". Additionally, we retrieve the list of textual descriptions associated with the human's clothing, if exist. We consider each text as a separate sentence.

Finally, we compile a list of sentences containing a caption prefix, an enumeration of the objects, the pairwise positional relations between objects, a scene description, and action and clothing descriptions for each human. We concatenate the sentences together to get a full dense caption of the image. A simplified pseudo-code for generating a caption is shown below:

\scriptsize{
\begin{lstlisting}[language=Python, caption=General Code for Caption Generation]
def sample_prompt(objects, seg_image, scene_name):
  statements = ["This scene contains "]

  # Objects
  objects_statement = ""
  for obj in objects:
    article = get_article(obj.obj_name[0])
    objects_statement += article + f" {obj.obj_name}, "
  objects_statement += f"and {len(humans)} humans."
  statements.append(objects_statement)

  # Scene statement
  scene_statement = "They are in "
  scene_article = get_article(scene_name)
  scene_description = get_description(scene_description)
  scene_statement += scene_article + scene_description
  statements.append(scene_statement)

  # Positional relations
  relations = []
  n = len(len(objects))
  for i in range(n):
    for j in range(i + 1, n):
      left, right = get_left_right(seg_image, i, j)
      front, back = get_front_back(i, j)
      relations.extend([
        left + " is to the left of " + right,
        right + " is to the right of " + left,
      ])
      relations.extend([
        front + " is in front of " + back,
        back + " is behind " + front,
      ])
  shuffle(relations)  
  statements.extend(relations)

  # Clothing and action
  for h in humans:
    # Action
    s_action = f"The {h.name} {h.action}."
    statements.append(s_action)

    # Clothing
    for s in h.clothe:
      s_clothe = f"The {h.name} {s.strip()}."
      statements.append(s_clothe)

  return " ".join(statements).strip()

\end{lstlisting}
}

\normalsize
Evidently, dense captions tend to be way too descriptive and hence noisy to be used fully in \vl{} training. Therefore, we add a sampling option where statements are sampled with certain probabilities following their weights. For example, instead of mentioning all pairwise positional relations, this option allows sampling a number of sentences from the positional relations category.
\newpage
\section{LLM-Based Caption Paraphrasing}\label{sec:sm:llms}
We additionally experiment with using the rule-based system to guide the use of large language models for caption generation / paraphrasing. Specifically, we adapt the deterministically-generated caption (as detailed in Section~\ref{sec:sm:grammar}) into a prompt for instruction-based text completion by replacing the prefix "This scene contains" (in the synthesized captions) to "Please describe a scene containing" and adding a suffix for text completion: "In this scene, we can see". We use the adapted texts as prompts for language models and generate text completions. We limit the generated texts to 150 tokens and use caption split averaging, as described in 
% Section~3.2
Section~\ref{sec:training}
% \if\sepappendix1{Section~3.2}
% \else{Section~\ref{sec:training}}
in the main paper. We experiment with the Bloomz 7.1B \cite{bloomz} and Flan-T5 XXL \cite{flant5} through Huggingface \cite{huggingface}.

Table \ref{tab:llm_extra} shows the performance of \ours{}CLIP trained using different caption generation mechanisms. We do not observe any significant performance gains when using the captions generated by openly avaialble language models that we tried over the rule-based system. This is indeed expected, as the captions generated by current open language models tend to repeat much of the content in the prompt, often correcting verb tenses or adding appropriate punctuation marks, which don't contribute to the semantic richness of the caption.

\begin{table}[h!]
\small
    \centering
    \begin{tabular}{l|ll|l}
            \toprule
            & \multicolumn{3}{c}{VL-Checklist}\\
            & Relation & Attribute & \textbf{Average}\\
            \midrule

            CLIP &  63.57 & 67.51 & 65.54 \\
            \midrule
            \ours{}CLIP with Rule-Based & 69.39 & 70.37	& \textbf{69.88} \\
            \ours{}CLIP with Bloomz & \textbf{69.66} & 69.69 &	69.68 \\
            \ours{}CLIP with Flan-T5 & 68.48 & \textbf{71.14} & 69.81 \\
            \bottomrule
    \end{tabular}
    
    \vspace{0.15in}
    \caption{VL-Checklist \cite{vlc} performance on variants of \ours{}CLIP fine-tuned on \ourdataset{} with captions generated using the rule-based system described in section 2 or using language models as described in section 3.}
    \label{tab:llm_extra}
\end{table}

However, we remark that additional work on using LLMs for caption generation could investigate more powerful language models, or the use of visual grounding for caption generation as an additional information source, to yield better paraphrasing / captions.
% \newpage
\section{Exploration into Synthetic Data Diversity}\label{sec:sm:diversity}
\begin{figure}[H]%
    \centering
        {\includegraphics[width=6.5cm]{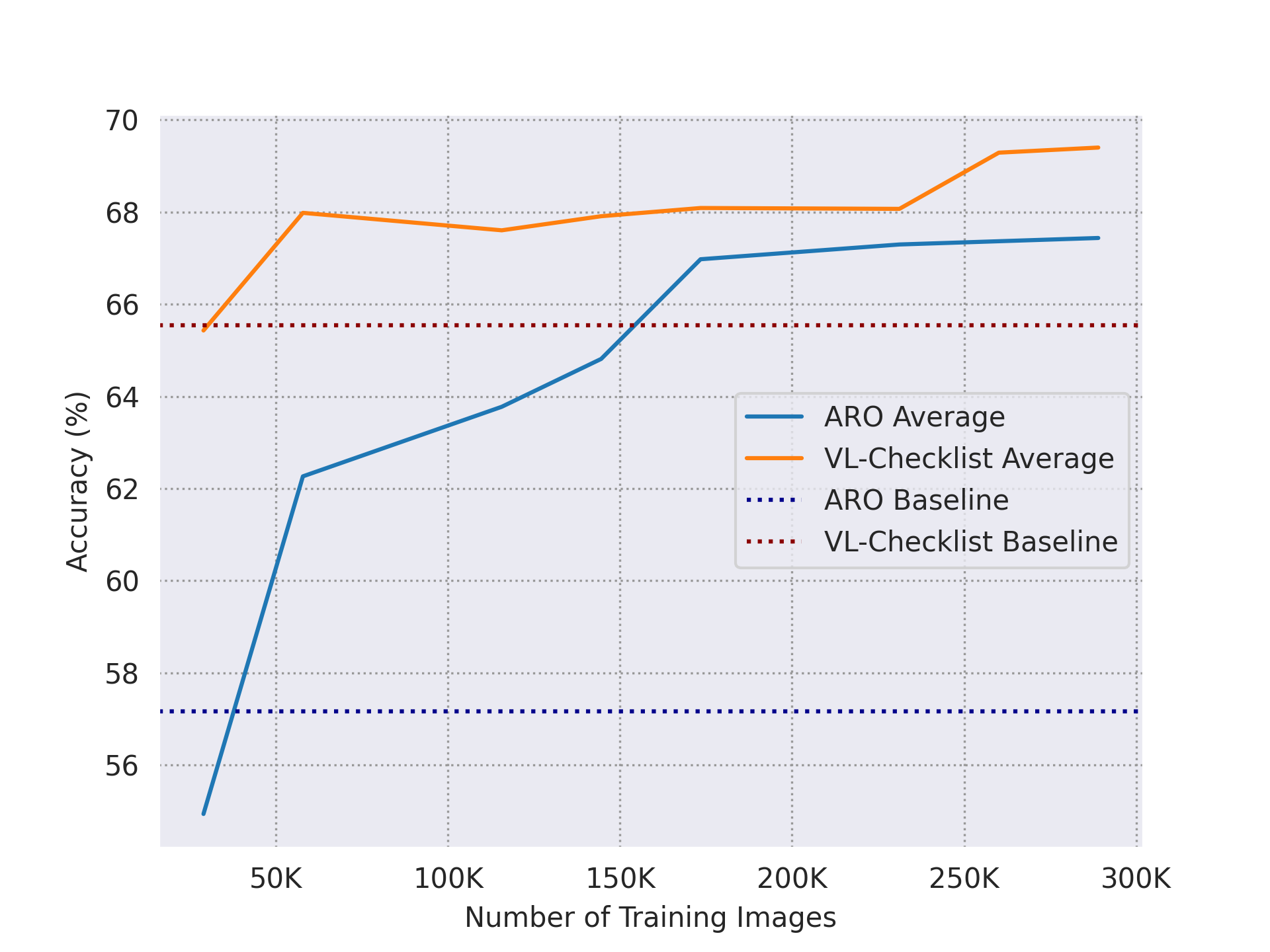}}%
        \;
        {\includegraphics[width=6.5cm]{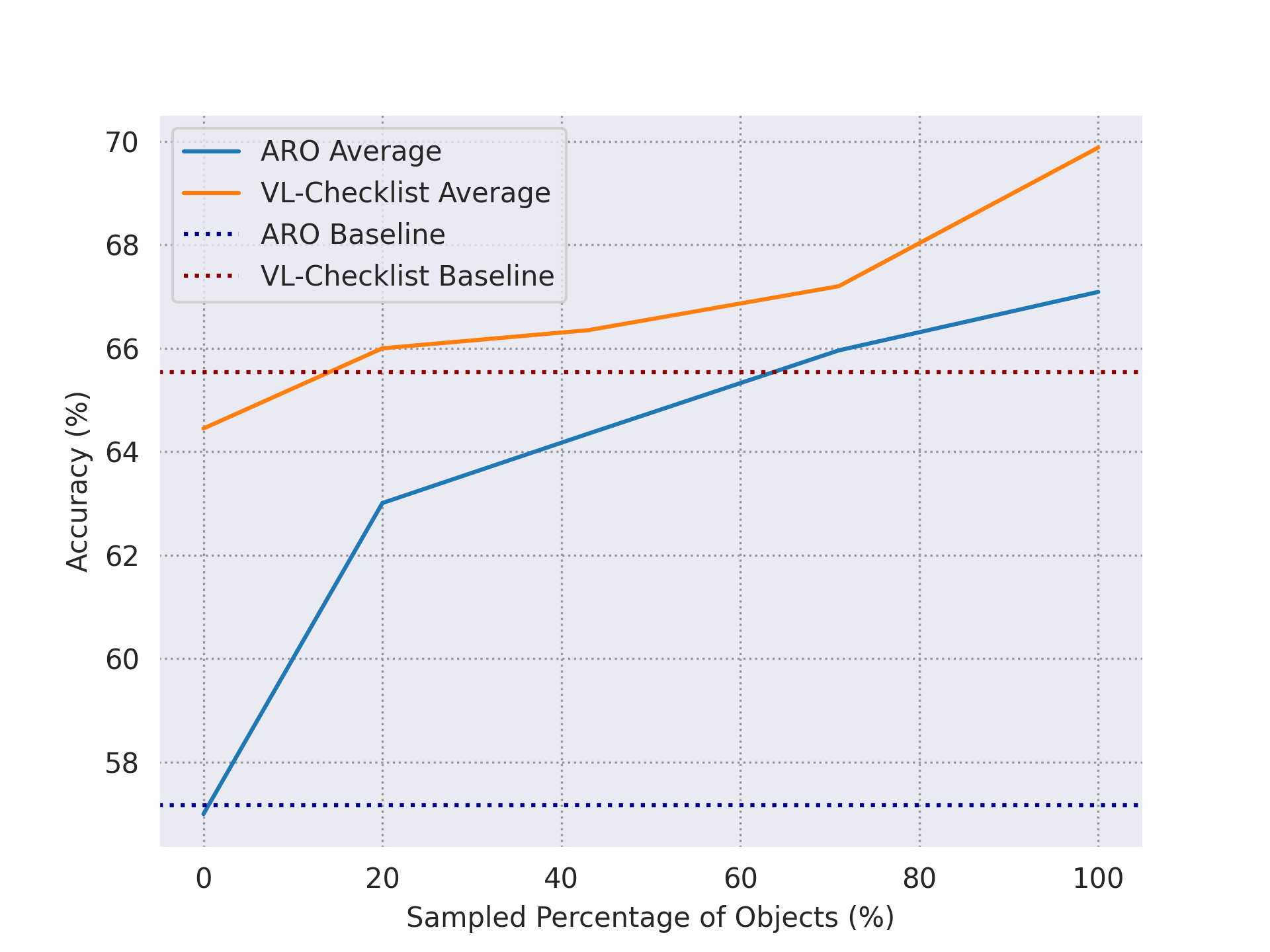}}%
    \caption{Exploration into Synthetic Data Diversity. (a) effect of adding more synthetic samples to \ourdataset{}; (b) effect of adding more 3D object models to \ourdataset{}. Comparing base CLIP and \ours{}CLIP performances on \vlchecklist{} and \ARO{} benchmarks.}
    \label{fig:diversity42}%
\end{figure}

As promised in Ablations 
% Section~4.4
Section~\ref{sec:ablations}
% \if\sepappendix1{Section~4.4}
% \else{Section~\ref{sec:ablations}}
of the main paper (lines 740-741 in "\ourdataset{} - number of models and number of samples") we include the effect of the number of synthetic samples and the number of object models used for \ourdataset{} generation analysis in Figure \ref{fig:diversity42}. These ablations were not included in the main paper due to lack of space. As we can see the performance is improving consistently, both with adding more synthetic images (Fig. \ref{fig:diversity42}a) and with adding more 3D models used for synthesis (Fig. \ref{fig:diversity42}b).

% ~\\
% ~\\
% ~\\
\newpage

\section{Some Qualitative Examples with Synthetic Humans}\label{sec:sm:qual}

\begin{figure}[h!]
    \centering
    \includegraphics[width=\linewidth]{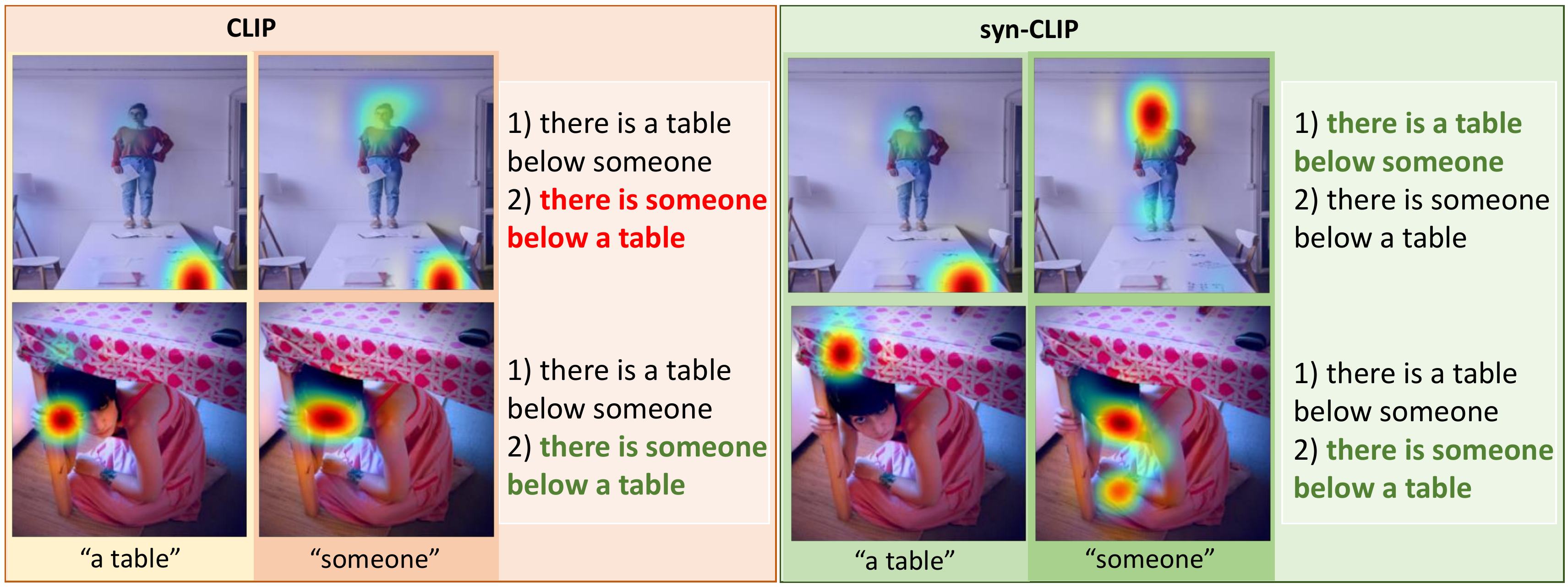}
    \caption{GradCAM on a Winoground sample. 
        \textit{left} - a CLIP model attending incorrectly to table regions for both \texttt{a table} and \texttt{someone} text queries, making a mistake in prediction (red text). \textit{right} - our syn-CLIP model correctly attends to the same making no mistakes in prediction with respect to the given inputs. Best viewed in color. } 
    \label{fig:grad_cam_examples}
\end{figure}

\begin{figure}[h!]
    \centering
    \includegraphics[width=\linewidth]{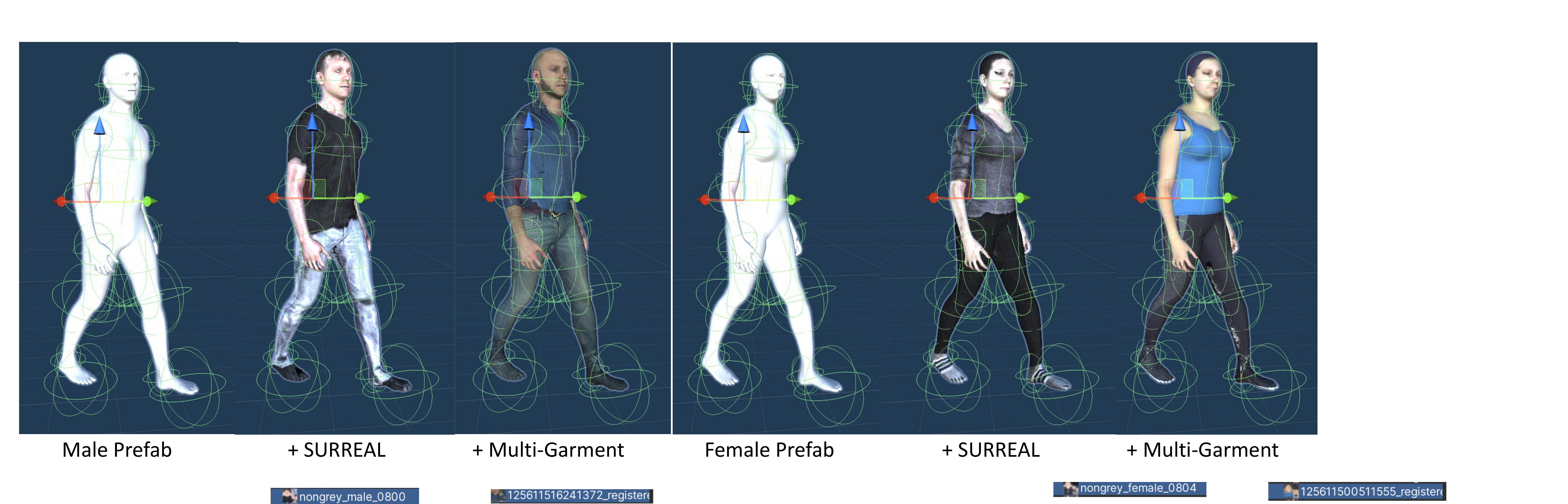}
    \caption{Digital humans. We show male and female samples of Unity Prefabs containing SMPL templates, a set of $514$ reusable 3D object assets available in \ourdataset{}. We add colliders to each model to allow interactions with other objects. } 
    \label{fig:sup_SMPLSamples}
\end{figure}

In this section, we first showcase
qualitative improvement in the compositional capabilities of CLIP after finetuning on SyViC using our proposed approach via GradCAM in Figure~\ref{fig:grad_cam_examples}. Next, we show textured SMPL samples in Figure~\ref{fig:sup_SMPLSamples}. 

Finally, we showcase some visual examples from \ourdataset{} along with the dense captions we generate describing human actions and detailed human-object interactions and relative position descriptions in the following pages. 

% \begin{figure*}[t!]
%     \centering
%     \includegraphics[width=0.6\linewidth]{Images/video_screenshots.pdf}
%     \caption{Screenshots of our video showcasing part of the image/text generation process. } 
%     \label{fig:sup_SMPLSamples2}
% \end{figure*}

\begin{figure*}[h]
    \centering
    \includegraphics[width=\linewidth]{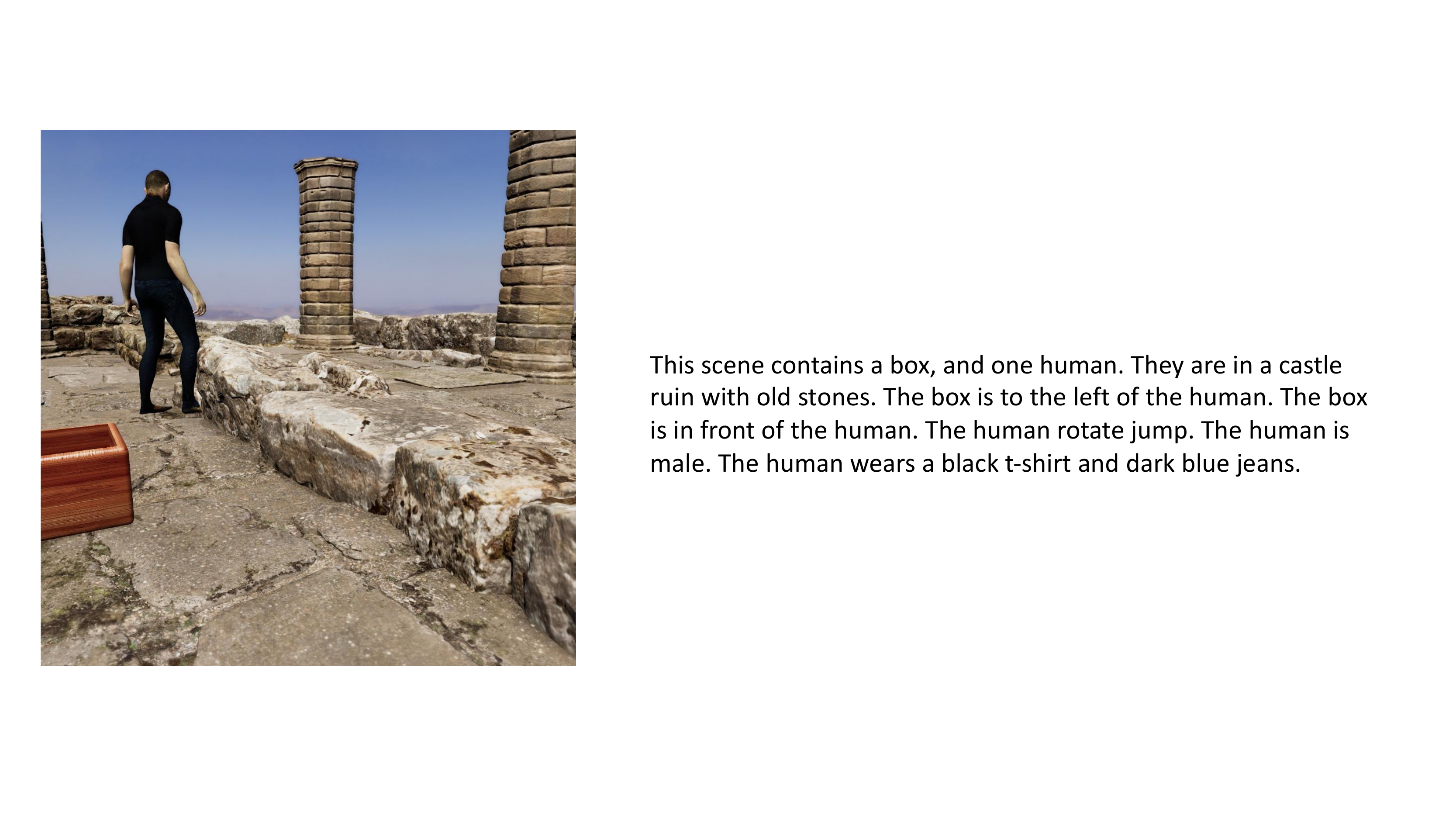}
    \label{fig:sup_qualitative1}
\end{figure*}

\begin{figure*}[h]
    \centering
    \includegraphics[width=\linewidth]{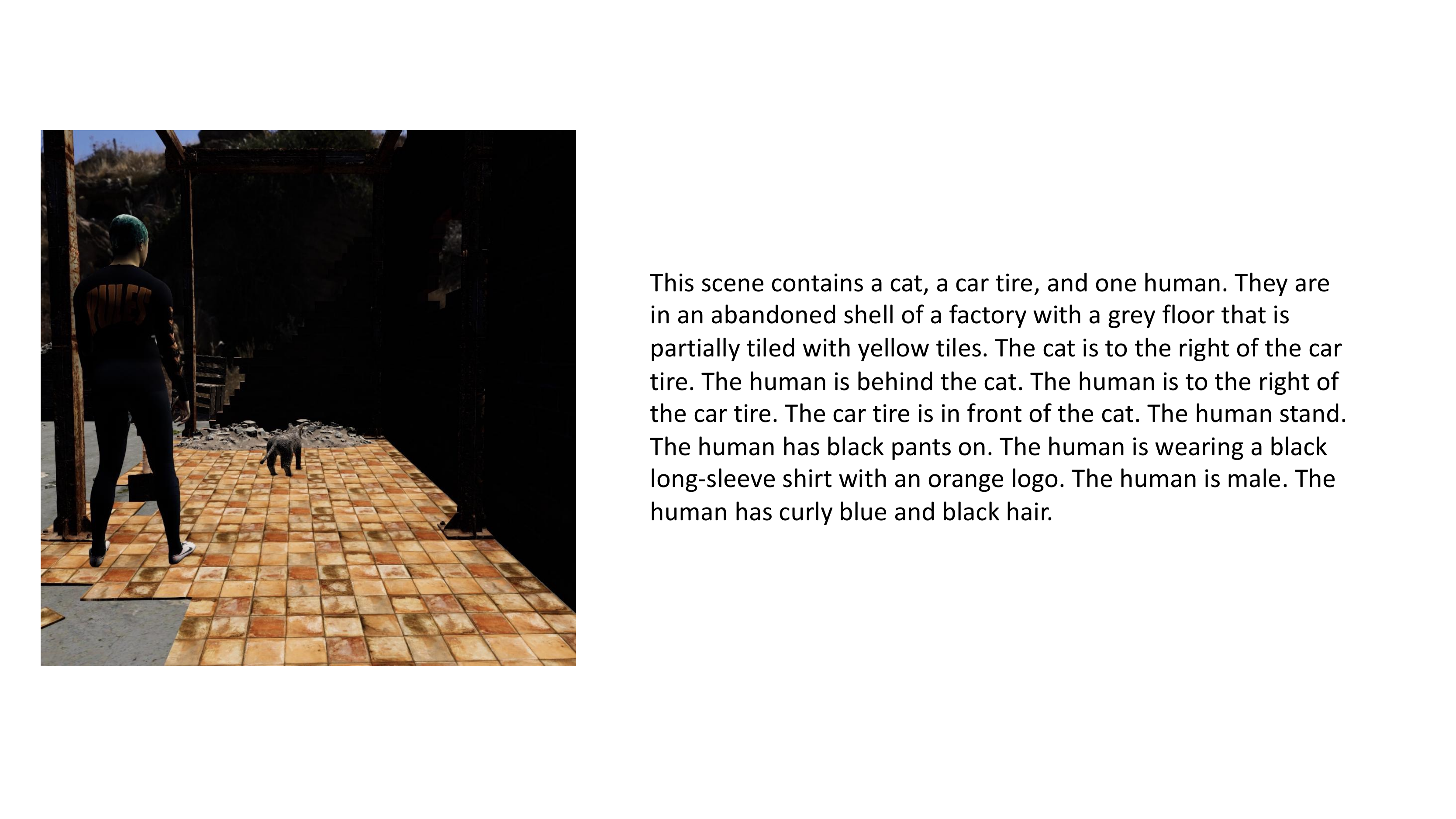}
    \label{fig:sup_qualitative2}
\end{figure*}

\begin{figure*}[h]
    \centering
    \includegraphics[width=\linewidth]{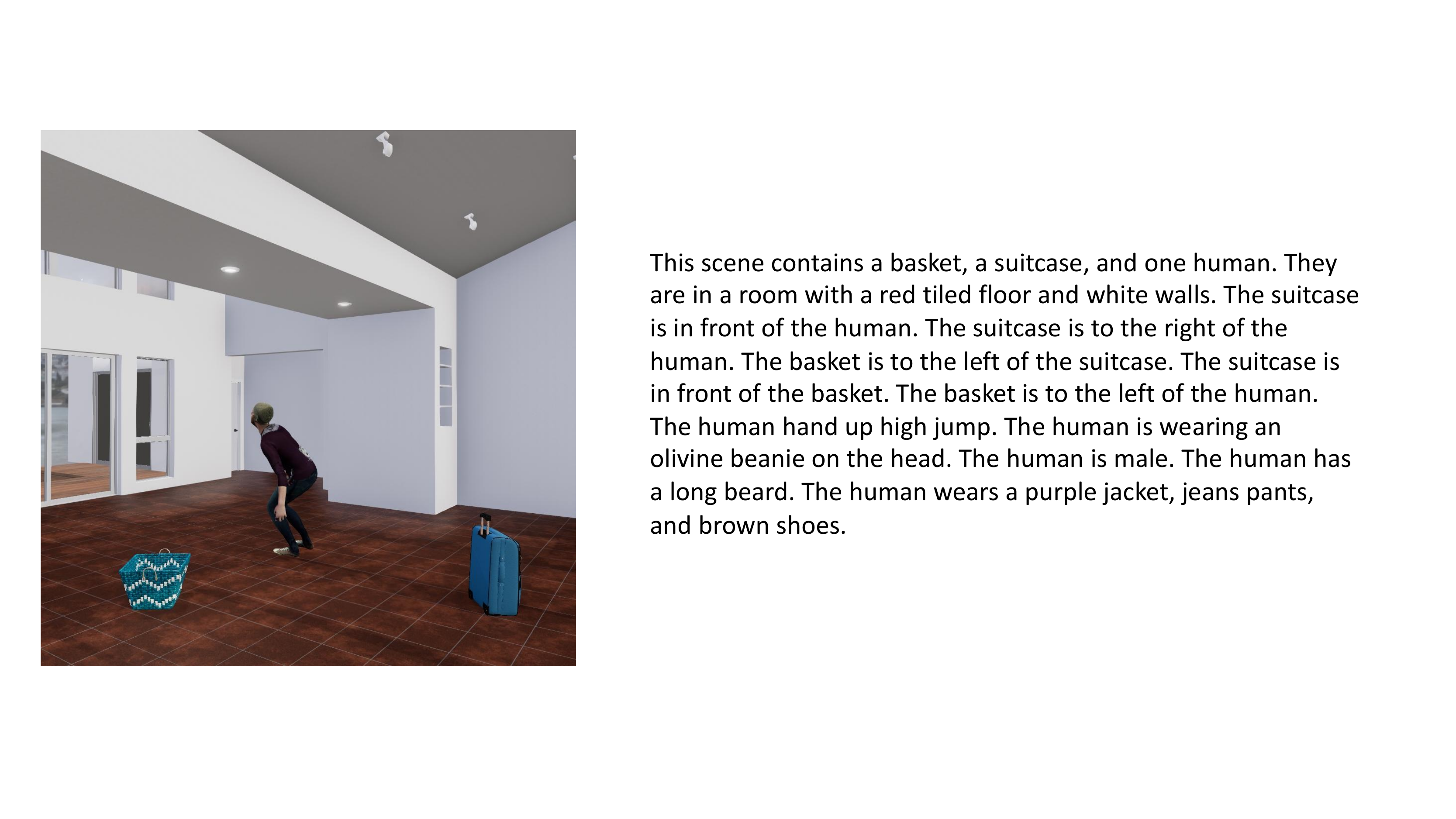}
    \label{fig:sup_qualitative3}
\end{figure*}

\begin{figure*}[t!]
    \centering
    \includegraphics[width=\linewidth]{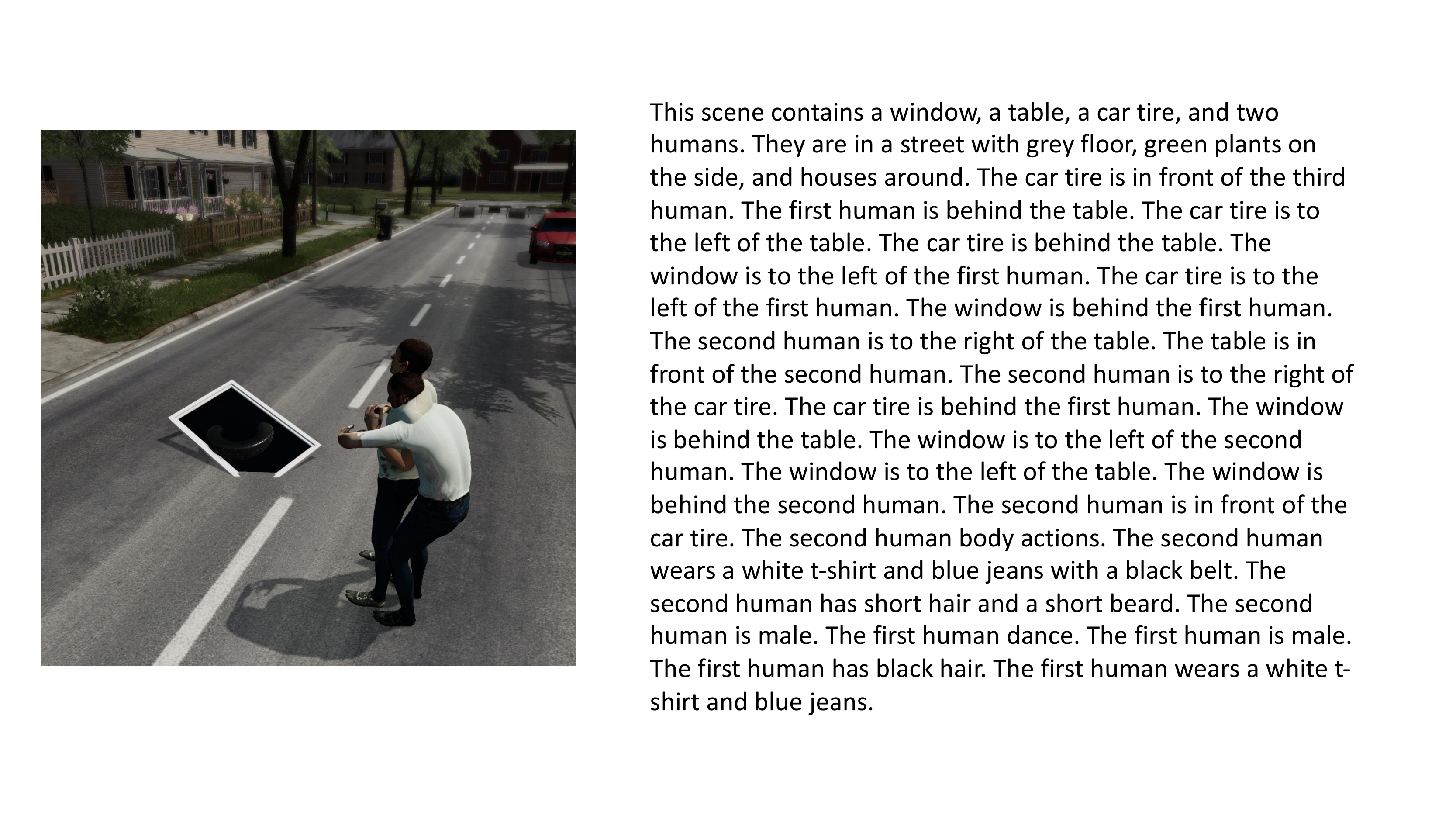}
    \label{fig:sup_qualitative4}
\end{figure*}

\begin{figure*}[t!]
    \centering
    \includegraphics[width=\linewidth]{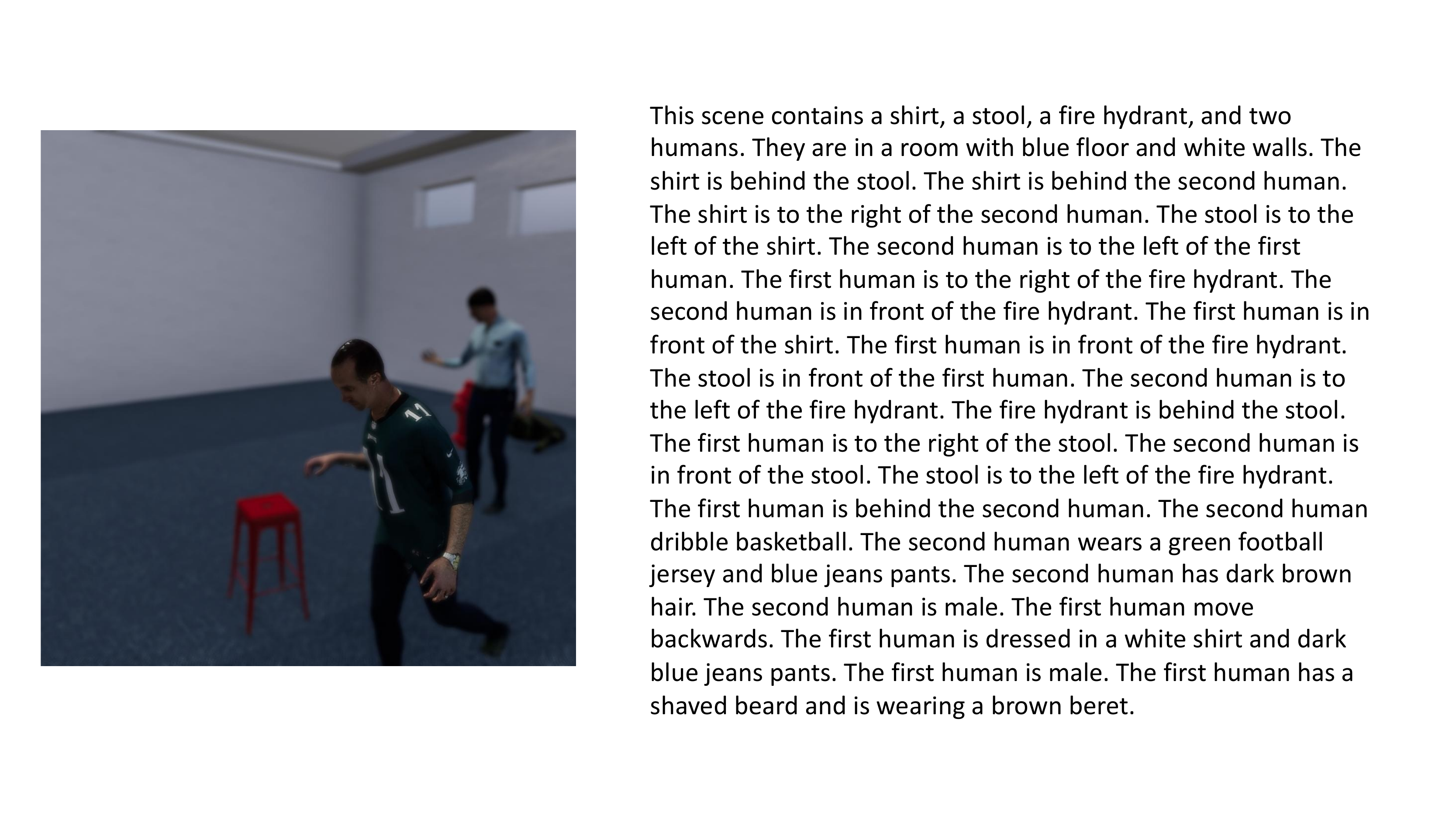}
    \label{fig:sup_qualitative5}
\end{figure*}

\begin{figure*}[t!]
    \centering
    \includegraphics[width=\linewidth]{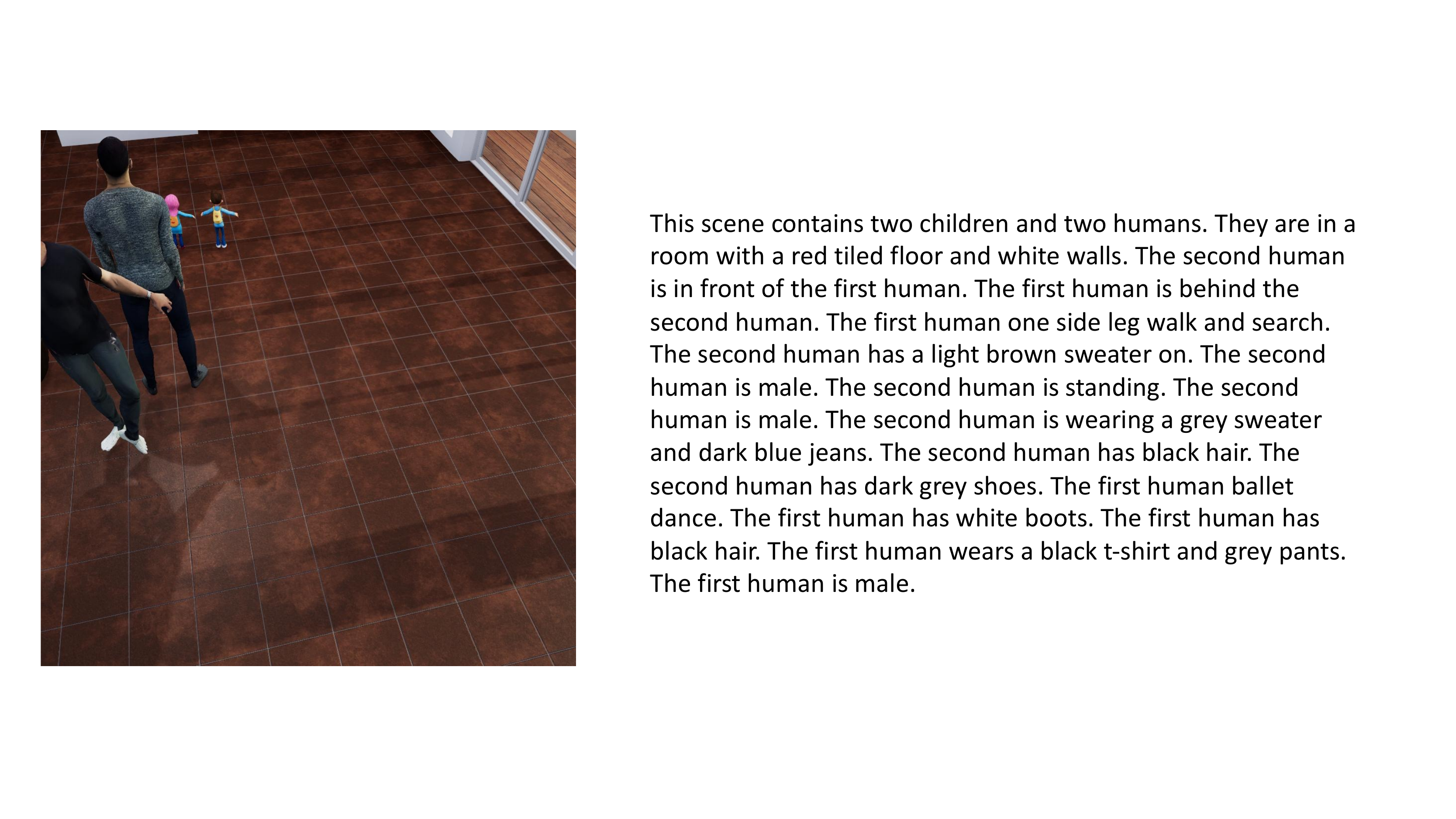}
    \label{fig:sup_qualitative6}
\end{figure*}

\begin{figure*}[t!]
    \centering
    \includegraphics[width=\linewidth]{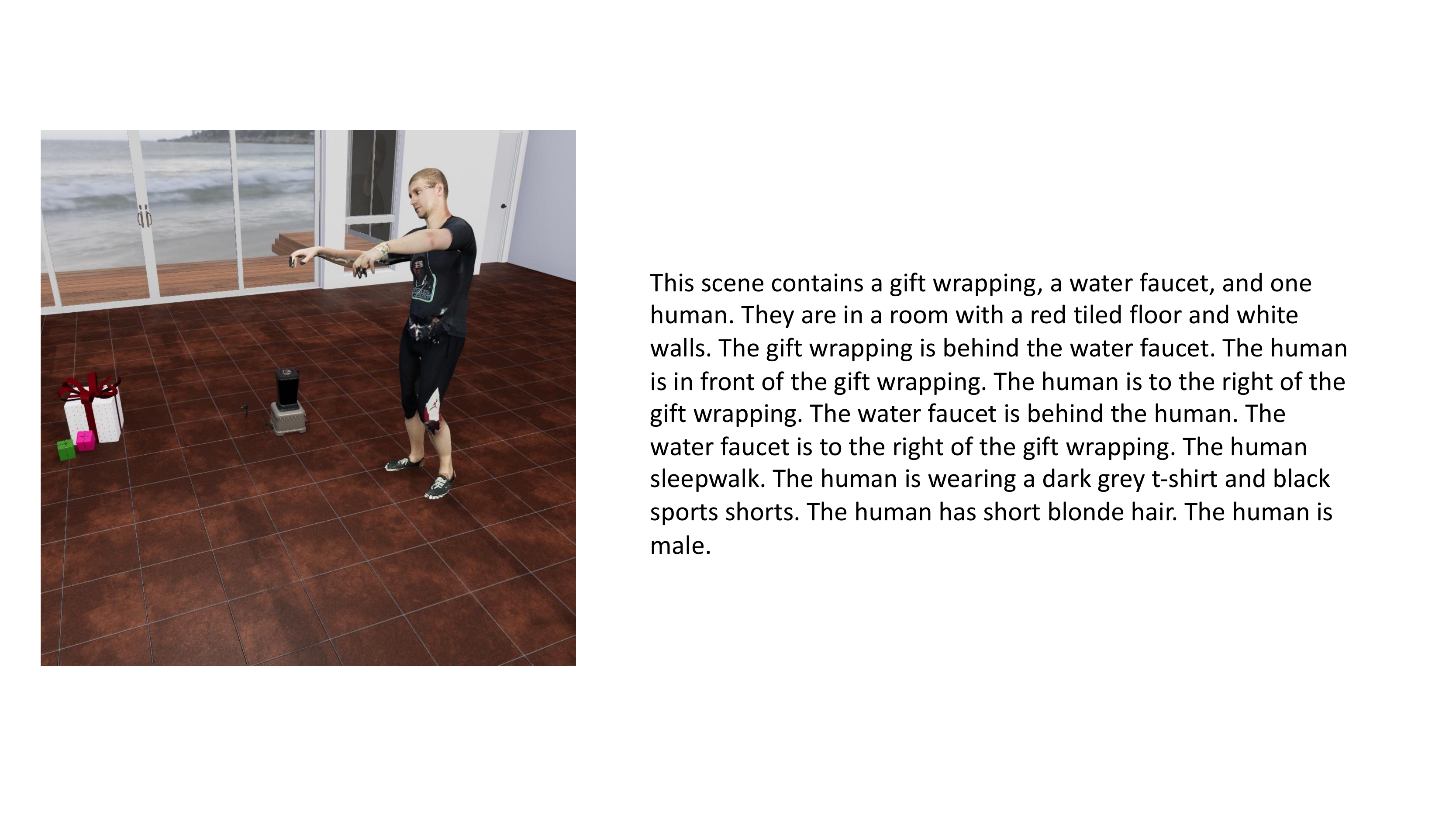}
    \label{fig:sup_qualitative7}
\end{figure*}

\begin{figure*}[t!]
    \centering
    \includegraphics[width=\linewidth]{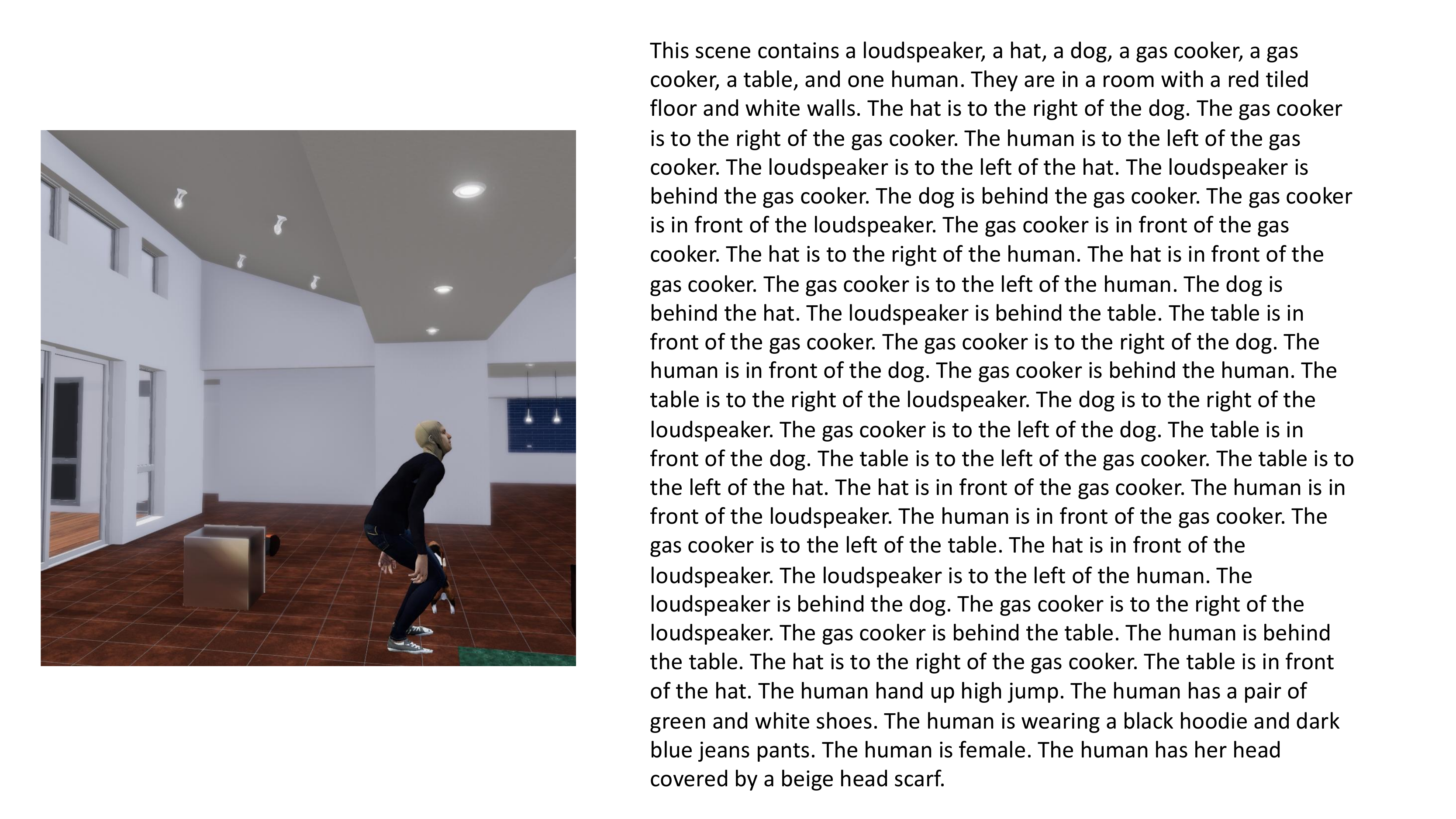}
    \label{fig:sup_qualitative8}
\end{figure*}

\end{document}